\documentclass{article}
\usepackage[utf8]{inputenc}

\usepackage[T1]{fontenc}
\usepackage{booktabs} 
\usepackage{algorithm}%
\usepackage{algpseudocode}%
\usepackage{hyperref}
\usepackage{amsmath}
\usepackage{titlesec}
\usepackage{xcolor}
\usepackage{xspace}
\usepackage{soul}
\newcommand{\ignore}[1]{}
\usepackage[textsize=tiny]{todonotes}
\usepackage{graphicx}
\usepackage{boldline}
\usepackage{lipsum}
\usepackage{graphicx}
\usepackage{subfig}
\usepackage{setspace}
\usepackage[export]{adjustbox}
\usepackage{adjustbox}
\usepackage{enumerate}
\usepackage{multicol}
\usepackage{multirow}
\usepackage{amsfonts}
\usepackage{colortbl}
\usepackage{float}
\usepackage{rotating}
\usepackage{setspace}
\usepackage{gensymb}	
\usepackage[official]{eurosym} 
\usepackage{siunitx}
\DeclareSIUnit{\EUR}{\text{\euro}}
\usepackage[flushleft]{threeparttable}
\usepackage{enumitem}
\usepackage{smartdiagram}
\usepackage{tabu,hhline,multirow}
\usepackage[justification=centering]{caption}
\usepackage[utf8]{inputenc}
\usepackage{fourier} 
\usepackage{array}
\usepackage{makecell}
\usepackage{url}
\usepackage{lineno, hyperref}
\usepackage{authblk}

\title{Optimization Study of Hydraulic Power Take-off System for an Ocean Wave Energy Converter}
\author[1]{Erfan Amini}
\author[1]{Hossein Mehdipour}
\author[2]{Emilio Faraggiana}
\author[1]{Danial Golbaz}
\author[3]{Sevda Mozaffari}
\author[2]{Giovanni Bracco}
\author[4]{Mehdi Neshat$^*$}
\date{\footnotesize *Corresponding author}

\affil[1]{\footnotesize School of Civil Engineering, College of Engineering, University of Tehran, Tehran. Iran, erfan.amini@ut.ac.ir; hossein.mehdipour@ut.ac.ir;  Dgolbaz@ut.ac.ir}
\affil[2]{\footnotesize Department of Mechanical and Aerospace Engineering, Polytechnic University of Turin, Torino, Italy. emilio.faraggiana@polito.it; giovanni.bracco@polito.it}
\affil[3]{\footnotesize Department of Water Engineering, Urmia University, Urmia. Iran. sevda.mzf@mail.com}
\affil[4]{\footnotesize Center for Artificial Intelligence Research and Optimization, Torrens University Australia, Brisbane, Australia. neshat.mehdi@gmail.com}

\begin{document}

\maketitle

\abstract{ Ocean wave renewable energy is fast becoming a key part of renewable energy industries over the recent decades. By developing wave energy converters as the main converter technology in this process, their power take-off (PTO) systems have been investigated. Adjusting PTO parameters is a challenging optimization problem because there is a complex and nonlinear relationship between these parameters and the absorbed power output. In this regard, this study aims to optimize the PTO system parameters of a point absorber wave energy converter in the wave scenario in Perth, on Western Australian coasts. The converter is numerically designed to oscillate against irregular and multi-dimensional waves and a sensitivity analysis for PTO settings is performed. Then, to find the optimal PTO system parameters which lead to the highest power output, ten optimization algorithms are incorporated to solve the non-linear problem, Including Nelder-Mead search method, Active-set method, Sequential quadratic Programming method (SQP), Multi-Verse Optimizer (MVO), and six modified combination of Genetic, Surrogate and fminsearch algorithms. After a feasibility landscape analysis, the optimization outcome is carried out and gives us the best answer in terms of PTO system settings. Finally, the investigation shows that the modified combinations of Genetic, Surrogate, and fminsearch algorithms can outperform the others in the studied wave scenario, as well as the interaction between PTO system variables.}


\section{Introduction}

Harvesting ocean waves energy has become more prominent since the oil crisis in 1973. Ocean energy has huge potential in providing energy especially for coastal communities, which made the governments, industries and engineers very attracted to this matter and resulted in more than 1000 patents worldwide \cite{haraguchi2020enhanced,bozzi2013modeling}. Wave energy has immense reserves, high power potential, higher power density than solar and wind energy, is geographically diverse and makes small interference in the environment \cite{yue2019dynamic,zhang2021evaluation}.

One of the key parts of a WEC is the power take-off system which converts the mechanical energy absorbed by the device into electrical energy \cite{zhang2015state,yue2019dynamic}.  Since the wave energy is fluctuating by nature, regular PTOs are not very effective at absorbing wave power [i31]. The development of hydraulic PTO systems has been a huge breakthrough in the field of ocean energy applications, due to the fast-frequency response, easy and high controllability, high efficiency, hydraulic overload protection, and adaptability to high power and low-frequency of the HPTO systems \cite{zhang2015state,yue2019dynamic,jusoh2021estimation}. These characteristics made HPTO the best PTO type for the point absorber wave energy devices, which can reach up to 90\% efficiency \cite{jusoh2021estimation}. \\
There has been an increasing number of studies about the HPTO application in different WECs in recent years \cite{galvan2020dynamic,penalba2018high,liu2018influence,cargo2014optimisation,brito2020experimental}, but the number of studies is still lower than the studies WECs with Linear PTO, and there are a lot of unexplored areas that need to be addressed in the future. Most of these studies investigated the performance and efficiency of the HPTO systems, but they did not consider the effect of HPTO model parameters. These parameters are really important because they affect the unit’s efficiency and power output. Only a small number of studies addressed this problem, especially the effect of simultaneous change in the HPTO parameters has been investigated in very few studies, for example in \cite{jusoh2020parameters,jusoh2021estimation}. That is why in this paper we investigate the effect of HPTO parameters, namely piston area, volume and pre-charged pressure of the low-pressure gas accumulator (LPA) and the volume of the high-pressure gas accumulator (HPA) simultaneously on the power output of the WEC. \\
Vantorre \textit{et al.} \cite{vantorre2004modelling} studied the influence of a few factors such as the WEC hull geometry, the external damping and additional inertia on a point absorber wave-energy device performance. In \cite{peiffer2011design}, the functioning of the WindWaveFloat concept, which includes a spherical shape WEC, the SWEDE, was analyzed. 
Yu \textit{et al.} \cite{yu2013reynolds} studied the performance of a floating point absorber utilizing a Computational Fluid Dynamics (CFD) method which is based on Reynolds-averaged Navier–Stokes (RANS). They found that the non-linear parameters, particularly viscous damping and wave overtopping, could reduce the absorbed energy of the floating point absorber notably, especially in bigger waves. In \cite{sjokvist2014optimization}, the effects of buoy’s radius and draft on the performance of a point absorber device have been studied utilizing a velocity ratio. The authors reported that there are optimal values for these parameters according to the generator damping of the device. \\
Clark \textit{et al.} \cite{clark2019towards} used the damage equivalent loads (DELs) on the power take-off system to investigate the influence of wave energy device shape and PTO reliability. They recommend not to take into account PTO-stroke constraint when evaluating the force-time series, because of inaccuracies of this method. Li \textit{et al.} \cite{li2020optimum} developed and investigated, a two-body self-react point absorber with 2 types of power take-off system, the mechanical motion rectifier (MMR) and non-MMR based. They found that the PTO inerter can increase the power output or move the peak frequency. Also, the device’s drag damping decreases the power capture in both scenarios. \\
In \cite{castro2020design} a two-body point absorber with an innovative electromechanical power take-off system was designed and analyzed. In \cite{ropero2020efficiency}, the performance of a point absorber was investigated using Smoothed Particle Hydrodynamics (SPH) method. They reported that keeping the device at water level puts the device under maximum loads, and in order to reduce it significantly, the buoy should be immersed at a specific depth. In \cite{haraguchi2020enhanced}, an innovative wave energy device with a tuned inertial mass (TIM) is designed, in which a supplementary tuning spring and a rotational inertial mass are affixed to the original system, which causes the oscillation of the damper to increase, which in turn can improve the WEC’s performance considerably. \\
In \cite{agyekum2021design}, a WEC emulator was developed, which can be used to evaluate a WEC’s performance, saving money and time. In \cite{karayaka2021investigations}, in order to decrease the cost of a wave energy converter PTO, the peak-to-average power (PTAP) ratio should be decreased while increasing the average absorbed energy by the device. Results showed that the most influential parameters on the PTAP ratio are gear ratio, crank radius, and generator properties. Zhang\textit{ et al.} \cite{zhang2021evaluation} investigated the performance of the bistable energy absorption mechanism in long-term ocean power capture. And found that, generally around truncated cone-shaped buoy with a bigger size and lower non-linear parameter can improve long-term energy absorption. \\
Numerical and metaheuristic algorithms have been utilized in WEC optimization in a lot of studies, here we briefly review a few of them. In \cite{faraggiana2018design}, the authors tried to minimize the LCOE of a multi-float WaveSub device using GA \& PSO. They considered a few decision variables, and found that the float-reactor separation is the key parameter to change.  Also, both algorithms had equally good performances. In \cite{amini2021comparative} the performance of 5 metaheuristics algorithms was compared in optimizing the power take-off system of a point absorber. Results showed that MVO had the best performance. \\
Noad\textit{ et al.} \cite{noad2015optimisation} investigated the performance of an OSWEC array. They determined the optimal parameters for a single OSWEC device using a numerical optimization method, and suggested that the gap length is the most important factor influencing the power output.  They also studied the optimal arrangements of a WEC array. \\
Babarit \textit{et al.} \cite{babarit2006shape} did a shape optimization study on a wave energy device, the objective was to maximize the power output and to minimize the cost simultaneously.  They utilized a 2 layer process to achieve this purpose, each layer using a different optimization algorithm, one of them was GA, and the other one was a gradient-based method. In \cite{jusoh2021estimation} researchers compared the performance of GA and NLPQL in optimizing the parameters of a hydraulic PTO system for a WEC. Both algorithms did improve the device’s performance significantly. \\
Plummer \textit{et al.} \cite{plummer2009investigating} developed an HPTO unit for a point absorber, and assessed the performance of the control strategies in both cases of with and without considering losses in the HPTO. Also, they showed that using a smaller motor and pump in case of smaller wave conditions can increase the device’s efficiency. In several recent studies \cite{amini2021comparative,quartier2018numerical,amini2021comparative2}, researchers studied different aspects of PTO systems utilizing optimization methods. For instance, A comparison has been done between two different WECs PTO systems, the Carnegie CETO 6 and the WaveRoller with a linear PTO and a hydraulic one. They compared the performance of three optimization strategies. Additionally, the effect of piston area on the power output of both devices was investigated. Further, \cite{neshat2021layout} developed a cutting edge swarm cooperative optimization framework for optimize power output of WEC farms.\\
Yu\textit{ et al.} \cite{yu2018numerical} developed an HPTO model for a two-body floating point absorber (FPA) to investigate the trade-off between the mean captured power and the power fluctuation using three power smoothing methods. These methods were able to lower the power fluctuation of the device notably. They also analyzed the effect of HPA Volume on the Power Output and its fluctuations. \\
In another study \cite{yue2019dynamic}, the effects of the HPA volume and pre-charge pressure, flow control valve and oil tank on the operational stability of an HPTO connected to a pendulum WEC were investigated. According to the findings, by increasing the accumulator pre-charge pressure or reducing its volume, the response speed for stabilizing the system state is decreased, but the stable state becomes more volatile in return. Finally, modifying these two parameters can improve the quality of the response speed, overshoot and operational stability of the system. \\
Cargo \textit{et al.} \cite{cargo2012determination} modeled a point absorber WEC using an HPTO with and without losses, to find whether the losses affect the parameters that maximize the power output. They could have altered 3 different parameters, but they only changed the motor displacement, because in practice this option is easier to manipulate, and found that changing this parameter can improve the performance in both cases. They continued their research in another study \cite{cargo2016strategies} where they examined a practical active tuning method for a heaving point absorber. They showed that over-simplification of the PTO dynamics in the design stage can lead to non-optimal designs. In this study, the influence of motor capacity and generator load on the absorbed power was analyzed. \\
Liu \textit{et al.} \cite{liu2018influence} investigated the effect of 7 HPTO system parameters, namely piston area, motor capacity, effective damping of the generator, mounting position of cylinder, rod-to-piston area ratio, volume and pre-charge pressure of the HPA on the power output of a two-raft-type WEC. They reported that the best values for the first 4 parameters are mostly dependent on the wave period. Also the rod area should be minimized while maintaining the necessary strength. And finally, higher values of HPA volume and pre-charge pressure don’t affect the capture width ratio, but lower values of these two parameters can indeed be consequential. \\
Zhang \textit{et al.} \cite{zhang2015state} investigated a state-dependent model of a hydraulic PTO for an inverse pendulum WEC. Depending on the design criteria, HPA volume and pre-charge pressure have different trends. In order to maintain constant pressure in the HPTO system, these values should preferably be higher, but lower values of HPA volume and pre-charge pressure are better for controlling the damping force applied to the hydraulic motor. In another study, Zhang\textit{ et al.} \cite{zhang2020parameter} showed that the performance of the WEC can be improved by altering the resistance load, the accumulator pre-charge pressure and the size of the opening of the regulating throttle valve. \\
Jusoh \textit{et al.} \cite{jusoh2020parameters} optimized the power output of a multi-point WEC using GA with 5 decision variables: the diameters of piston and rod, accumulator’s volume and pre-charge pressure and motor displacement. They were able to reduce the size of all of the 5 parameters up to 38\%, which in turn reduced the operational cost. They also reduced the PTO force applied to device, which makes it possible for the WECs to be employed in smaller wave conditions. The authors continued their work on the same WEC and added two more parameters to their power optimization study \cite{jusoh2021estimation} to make a total of seven decision variables, namely LPA volume capacity and pre-charge pressure. They also used NLPQL (Non-Linear Programming by Quadratic Lagrangian) alongside GA. The latter algorithm had slightly better results with performance improvements up to 97\%. \\
According to the spotted gap in the literature, this paper attempts to find the optimal parameters of an HPTO unit of a point absorber in a case study of western Australian coasts in order to maximize the output power of the WEC. Different optimization algorithms are used for this purpose, and their performances are compared at the end of the simulations. Moreover, a modified combination of Genetic, Surrogate and fminsearch algorithms is developed. Finally, the simultaneous interaction between different system settings has been investigated. Finally, the simultaneous interaction between different system settings has been investigated.  \\
In this regard, section 2 explains the research method considering wave resources, wave energy converter simulation and governing equations. It is followed by the PTO dynamic model and optimization approaches including numerical and metaheuristics methods. Then, section 3 represents the feasibility study together with the sensitivity analysis. Next, the optimization convergence for different algorithms is presented in section 4. Afterward, section 5 provides a comparative assessment of the optimization algorithms. Finally, the conclusion is given in section 6. \\

\section{Research Method} 
\label{Research Method}
In this section, we cover the methodology of the research. It begins by describing the wave resource, including the wave scatter data and power matrix. It follows by wave energy conversion simulation details including the hydrodynamic governing equations. Then the PTO dynamic model details are explained, and its mechanical setup is presented. After that, we go through the optimization algorithms' specifications that we used in this study, followed by a detailed representation of our modified combined optimization algorithm.

\subsection{Wave Resource}
\label{Wave Resource}

\begin{figure}[htb]
    \centering
    \captionsetup{justification=centering}
    \includegraphics[width=1\linewidth]{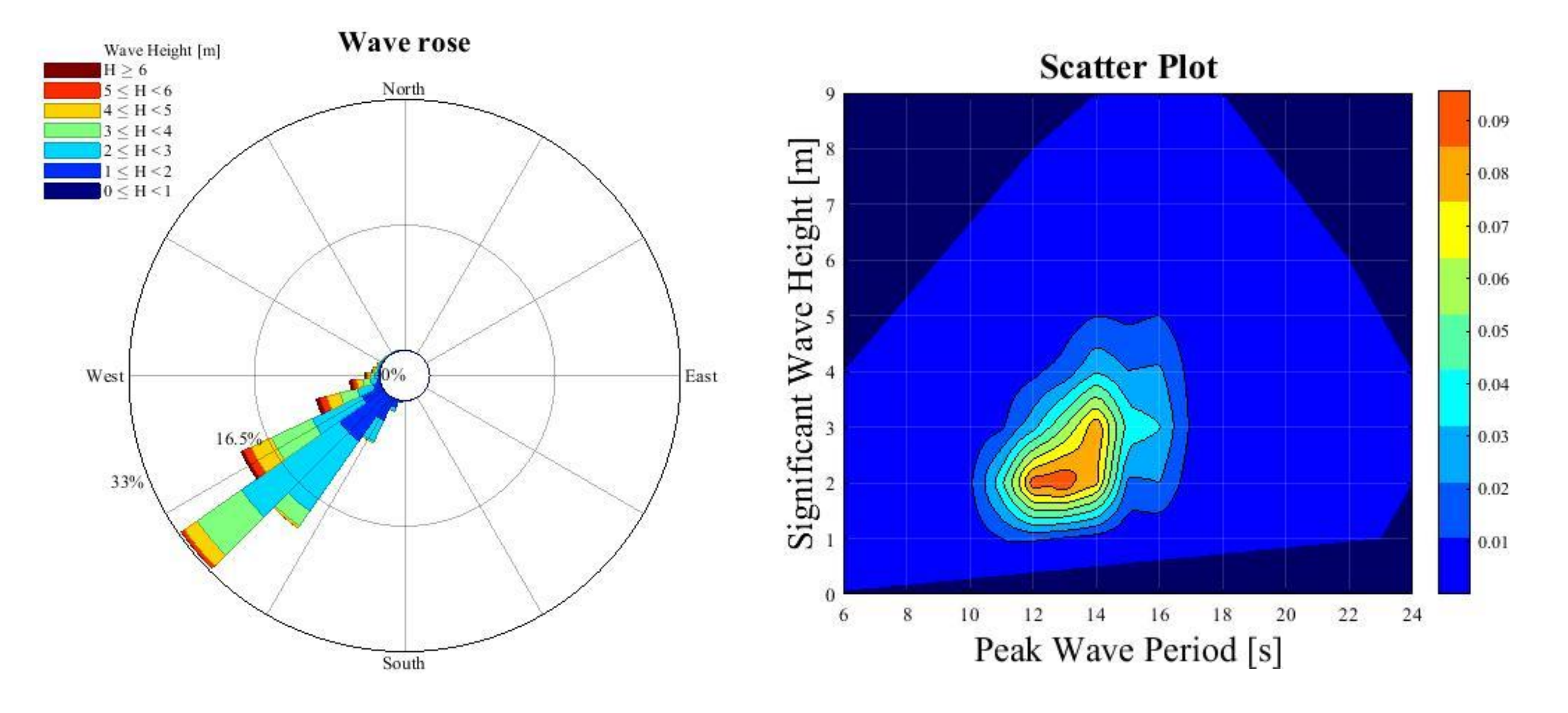}
    \caption{Wave rose of travelled waves in Perth (left), and wave scatter data (right) }
    \label{fig:WaveRoseScatter}
\end{figure}

\begin{figure}[htb]
    \centering
    \captionsetup{justification=centering}
    \includegraphics[width=0.6\linewidth]{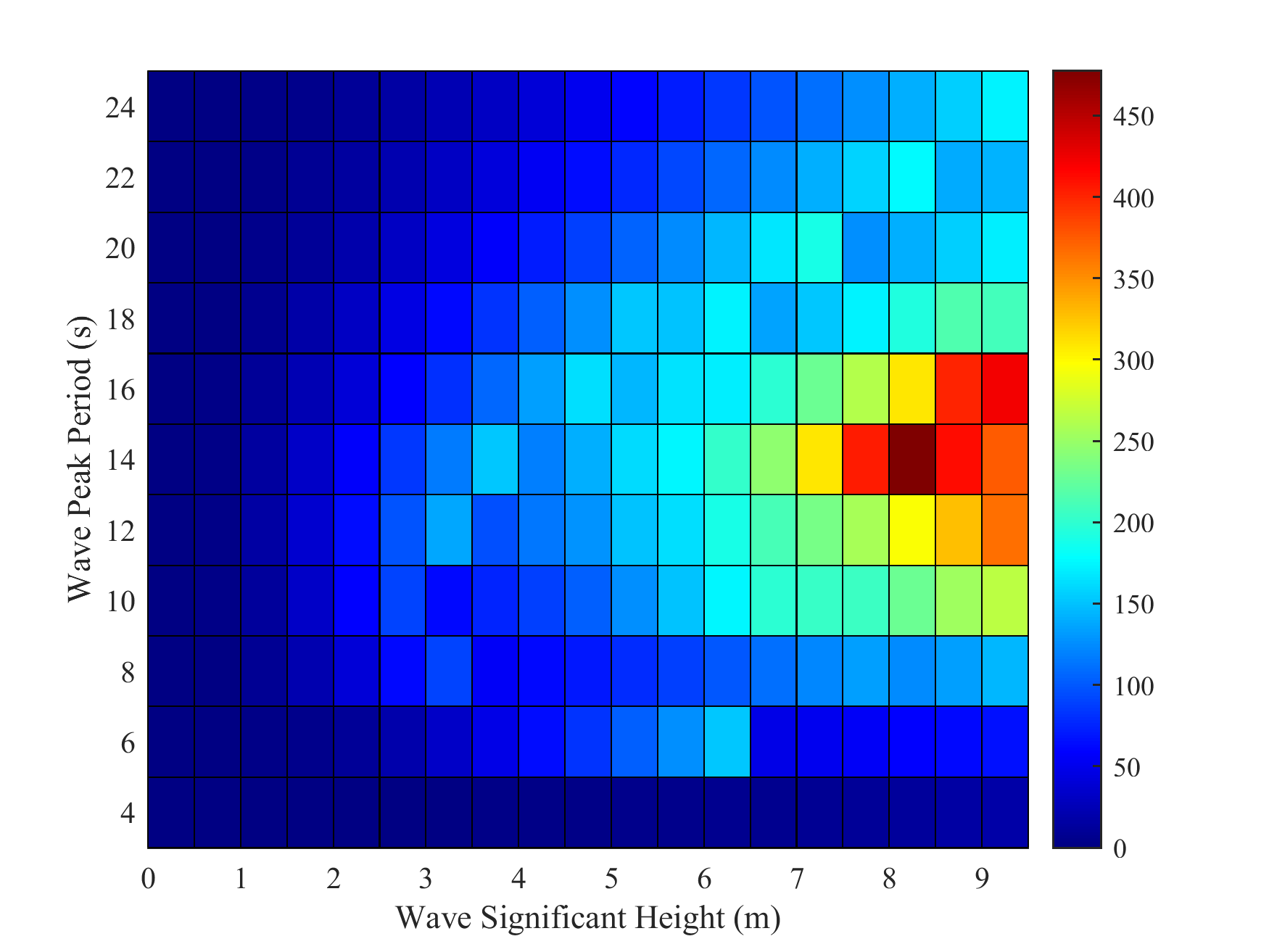}
    \caption{Power Matrix of waves in Perth }
    \label{fig:PowerMatrix}
\end{figure}

Looking back to the recent literature on potential wave energy in the Australian coasts, Perth has been identified as one of the most promising areas for the installation of such systems\cite{neshat2020new}. Accordingly, a wave data-set including the time history of waves' significant heights, periods, and directions is incorporated from the Australian Wave Atlas \cite{australian_renewable_energy_agency_2018}, and according to the literature, in this study, the Pierson- Moskowitz (PM) spectrum with a Significant Wave Height of 4.06 m and a Peak Wave Period of 13.65 s has been used to model the waves of this area. Figure \ref{fig:WaveRoseScatter} reveals the directional wave rose diagram and scatter data in Perth. For instance, the highest potential probability of occurrence is related to waves with a significant height of 2 meters, and a period of roughly 13 seconds. The power matrix of this area is presented in Figure \ref{fig:PowerMatrix}.

\subsection{WEC Simulation} 
\label{WEC Simulation}
WEC-Sim is a code utilized in this study and created by Sandia National Laboratories (SNL) and the National Renewable Energy Laboratory (NREL). Following that, time-domain modeling was complemented in MATLAB software, which is based on a multibody dynamic solver \cite{mathworks}. The wave energy converter requires hydrodynamic input to describe power output results, wec-sim imports are calculated by a frequency domain and pre-prepared by the BEMIO function.
In terms of the boundary element method's source program, NEMOH is utilized. The BEM is obtained by linear free surface potential flow theory and it generated hydrostatic and hydrodynamic coefficients. There are some contributors to relevant coefficients including mass, wave excitation term, hydrodynamic damping, and hydrostatic stiffness, given that those will be utilized by WEC-Sim. Considering the effect of these coefficients, relevant input data to waves, oscillating body, constraints, and the type of PTO are required by WEC-Sim. Providing a brief overview of how NEMOH determines the mentioned coefficients. The BEMIO functions take the data from NEMOH as input and calculate output non-dimensionalized hydrodynamic coefficients, as well as excitation and radiation impulse response functions \cite{wec123}.
By using NEMOH, we made the following assumptions \cite{babarit2015theoretical,de2009hydrodynamic}.
(i) The fluid is incompressible,
(ii) The fluid is inviscid,
(iii) The fluid is irrotational,
(iv) Both the waves and the oscillating body have small amplitudes,
(v) The seabed is flat, so the depth of water $d_w$ is supposed to be constant.
By discretizing the wetted surface in flat panels on which the velocity potential is intended to remain constant, the open-source BEM solver NEMOH calculates a solution for the velocity potential. More information on how NEMOH solves this boundary value problem may be found in \cite{babarit2015theoretical}. As indicated in Equation 1, WEC-Sim solves WEC's governing equations of motion using the Cummins in 6 degrees of freedom (DOF). 
\begin{equation}
F_e(t)-\int_{-\infty}^{t}\mathfrak{f}_r(t-\mathcal{T})\dot X(\mathcal{T}) d\mathcal{T}+ F_{hs}(X)+ F_m(X,\dot X)+ F_{PTO}(\dot X)-F_D(\dot X)=(m + A(\infty))\ddot{X}
\end{equation}
Where $F_e$ is the wave excitation force, the convolution integral is the wave radiation force, $F_{hs}$ is the hydrostatic restoring force, $F_m$ is the mooring force, $F_{PTO}$ is the PTO force, $F_D$ is the drag force, m is the WEC mass and $A_{\infty}$ is the added mass at infinite wave frequency \cite{ruehl2014preliminary}.
Reference Model 3 (RM3) is one of the WEC models developed by the U.S. Department of Energy\cite{sandia1}. And the design load elevation framework was demonstrated with this reference model\cite{yu2015experimental}.RM3 is a heavy two-body point absorber made up of a float and spar, whose full-scale dimensions and mass properties are illustrated in figure (2) and table(1) RM3 model. In reaction to incident waves, the WEC can move freely in all 6 DOF. In terms of the heave direction, power is predominantly captured. As a result of previous DOE-funded projects, there is a strong correlation between numerically design characterized and experimentally. Additionally, it has basic operating principles and is reflective of the wave energy industry's WECs. The total simulation time for cases simulated in this study is 400 s and the ramp time is 100 s.

\begin{table}[htp]
\caption{RM3 mass properties}
    \centering
    \begin{tabular}{|cccccc|}
    \hline
         \toprule    
        \multicolumn{4}{c}{Spar-plate Full scale Properties}  \\
     \hline
     Axis &\multicolumn{3}{c}{ Moment of Inertia [kg-$m^2$]}&  Mass [tonne] & CG [m]\\
    \hline
    X & 9.44E+07    & 0.00E+00  & 0.00E+00 & 878.30 & 0.00 \\
     \hline
    Y & 0.00E+00    & 9.44E+07   & 2.18E+05 & 878.30 & 0.00\\
     \hline
    Z& 0.00E+00    & 2.18E+05    & 2.85E+07 & 878.30 & -21.29\\
    \cline{1-6}
         \multicolumn{4}{c}{ Float Full scale Properties}  \\
     \hline
    X & 2.09E+07    & 0.00E+00  & 0.00E+00 & 727.01 & 0.00 \\
     \hline
    Y & 0.00E+00    & 2.13E+07   & 4.30E+03 & 727.01 & 0.00\\
     \hline
    Z & 0.00E+00    & 4.30E+03    & 3.71E+07 & 727.01 & -0.72\\
     \cline{1-6}
    \toprule
     \end{tabular}
    \end{table}
    
 \begin{figure}[htb]
    \centering
    \captionsetup{justification=centering}
    \includegraphics[width=1\linewidth]{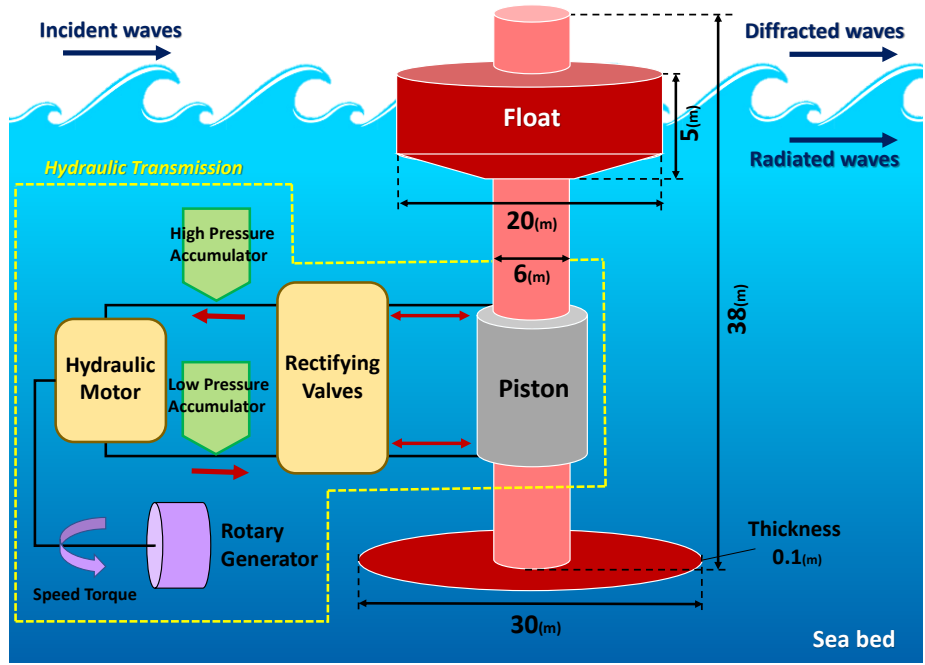}
    \caption{An illustrative look of the studied wave energy converter geometric dimensions, with a hydraulic power take-off system. }
    \label{OSWEC}
\end{figure}
\subsection{PTO Dynamic Model} 
\label{PTO Dynamic Model}
\begin{figure}[htb]
    \centering
    \captionsetup{justification=centering}
    \includegraphics[width=0.5\linewidth]{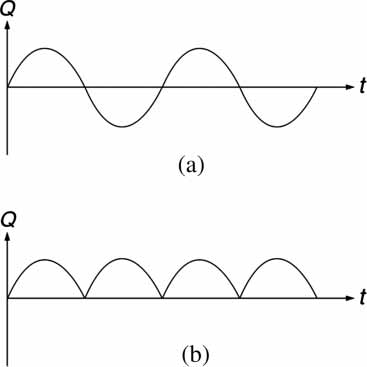}
    \caption{Flow-time representations for hydraulic WEC \cite{drew2009review}}
       \label{Flow-time}
\end{figure}

Most point absorbers capture energy based on body oscillations caused by wave interaction. In such cases,
conventional rotary electrical machines are not directly compatible. In most cases, Hydraulic Power Take-Off (HPTO) systems are chosen to deal with this problem by linking the WEC to an electrical generator, since they are well suited to capture energy when dealing with immense forces at low frequencies \cite{pecher2017handbook}. \\
The hydraulic cylinder's rod is forced up and down by the floating heaving buoy, with respect to an actuator \cite{pecher2017handbook}.
The first HPTO element is a double-acting hydraulic piston pump, which directly turns the linear kinetic energy of the heaving buoy into a pressurized, bi-directional fluid flow \cite{so2015development}. \\
The piston chamber is linked to a series of rectifying valves (4 valves) via upper and lower terminals. This mechanism rectifies the flow, meaning it converts the bi-directional fluid flow into a unidirectional flow and passes the fluid on to the rest of the HPTO system \cite{so2015development}.
One of the valve's roles is to deliver fluid to the high-pressure side of the system where it is stored in the High-Pressure gas Accumulator \cite{so2015development}.
 If the incident waves are almost sinusoidal, then the flow from one part of the cylinder can be represented by Figure\ref{Flow-time}(a). Rectification valves transform the fluid flow into a unidirectional flow that is depicted in Figure\ref{Flow-time}(b). Then the HPA would smooth this flow and send it to the hydraulic motor \cite{drew2009review}. 

The role of HPA is to smoothen the flow across the motor by either providing or accumulating hydraulic energy when needed, in other words, it reduces the hydraulic fluctuation in the system \cite{casey2013modeling}.
The smoothing effect increases with the accumulator volume and working pressure \cite{antonio2008phase}. \\
The hydraulic fluid power is converted into rotational energy by a variable displacement motor. The motor's shaft is connected directly (no gearboxes) to the shaft of the generator, causing it to revolve and generate electricity. The hydraulic motor was selected to meet the torque and speed required circumstances \cite{so2015development}. Since waves generate large loads, low velocity, oftentimes radial piston motor is used as it is well suited for these circumstances \cite{pecher2017handbook}.

The generator provides a gear speed ratio that translates the slow movement of the heaving buoy (WEC) to higher rotational speeds in the generator \cite{yu2018numerical}.
Electrical power is less than mechanical power because the generator block in the Simulink is modeled such that the rotational speed and torque of the electrical generator determine the efficiency of output electrical power, meaning there are some losses in the generator \cite{so2015development}.
The generator in the Simulink is modeled as simple rotational inertia with a lookup table containing speed, torque, and efficiency, based on a typical large industrial asynchronous generator \cite{so2015pto}.

Next, the fluid enters the low-pressure side where the LPA provides a pressurized reservoir. Whenever the hydraulic piston pump needs, it can draw fluid from the reservoir (LPA), to complete the circuit \cite{so2015development}.
The LPA is used to re-energize the hydraulic fluid \cite{yu2018numerical}, it also supplies a small boost in the pressure to reduce the risk of cavitation on the low-pressure side \cite{drew2009review}. 
 
 It is assumed that the hydraulic piston pump is directly coupled to the heaving buoy, therefore the velocity of the cylinder is the same as the buoy ($v$) \cite{casey2013modeling}, which is the relative velocity between the float and the spar \cite{so2015development}, and the output force can be determined using the pressure values of the upper and lower parts of the piston \cite{quartier2018numerical}: \\

\begin{equation}\label{FPTO}
 F_{PTO} = (-sign(v)).(P_{high} - P_{low}) A_p 
\end{equation}

with the $A_p$ the piston area, $P_{high}$ and $P_{low}$ the pressure in the upper section and lower section of the piston. Also, the total mean absorbed power of the HPTO is calculated by

\begin{equation}
P_{abs} = F_{PTO}.v = (sign(v).\Delta p.A_p).v 
\end{equation}

The next parameter is the volume flow $(Q_{piston})$  caused by the piston movement that can be calculated \cite{quartier2018numerical}: \\

\begin{equation}
Q_{piston} = A_p v
\end{equation}
the volume flow into the nitrogen-charged high-pressure accumulator (HPA) and low-pressure accumulator (LPA) can be determined this way. They have the same absolute value, but opposite signs. The rectifying valves guarantee that further in the hydraulic PTO, flow is unidirectional: fluid flows from the valves towards the HPA and from the LPA towards the valves \cite{quartier2018numerical}. \\
The accumulator's job is to smoothen the fluid flow to the hydraulic motor, to dump the undesired power output peaks \cite{quartier2018numerical}. \\
The incoming volume flow into the HPA ($Q_{in}$) is the sum of $Q_{piston}$ and $Q_{motor}$ - see further. Origin of $Q_{motor}$ is the hydraulic motor. In order to calculate the volume of fluid that flows into the HPA in time ($dt$), this volume is integrated \cite{quartier2018numerical}: \\
\begin{equation}
V_{in} = Q_{in}.dt
\end{equation}
Next, the pressure inside the HPA ($p_{high}$) is determined according to an isentropic process and is calculated per time step \cite{quartier2018numerical}: \\

\begin{equation}
p_{high} = \frac{p_{pre-charge}}{(1-\frac{V_{in}}{V_0})^{1.4}}
\end{equation}
The low pressure in the LPA can be calculated in a similar way \cite{quartier2018numerical}. \\
The angular velocity of the hydraulic motor ($\dot{\omega_m}$) can be determined by integrating the following equation \cite{quartier2018numerical}:
\begin{equation}
\dot{\omega_m} = \frac{((P_h - P_l).\alpha D - T_g - T_f)}{I_{mg}}
\end{equation}
where $T_g$ is the generator torque, $T_f$ the torque due to the friction, $\alpha$ the swashplate angle, $D$ the nominal motor displacement, and $I_{mg}$ the total moment of inertia of the motor/generator. Finally, the volume flow originating from the hydraulic motor ($Q_{motor}$) is given by \cite{quartier2018numerical}: \\

\begin{equation}
Q_{motor} = \omega_m \alpha D
\end{equation}
The generator damping can be calculated by:

\begin{equation}\label{Eq:Cgen}
C_{gen} = (\Delta P.\alpha D). (\frac{\omega_M}{\omega_{desired}}).(\frac{1}{1.05})
\end{equation}where $C_{gen}$ represents the generator damping, $\Delta$P the pressure difference between the HPA and LPA, $\alpha$D the hydraulic motor volume, $\omega_{desired}$ the desired speed of the rotary generator (150 rad/s) and 1.05 accounts for the mechanical efficiency. Also the term $(\frac{\omega_M}{\omega_{desired}})$  represents the volumetric efficiency \cite{Anu:2021}.

Finally, it is notable that our case uses a variable hydraulic motor displacement, and its value can be determined by:
\begin{equation}
\begin{cases}
    
if ~~4 MPa  < \Delta P <  15 MPa  \quad \rightarrow\quad    \quad   \alpha D  = 2.67e10^{-11}.\Delta P -  8.52e10^{-5} (m^3) \\

else \quad \quad \quad \quad \quad \quad \quad\quad \quad\quad \quad   \rightarrow   \quad\quad  \alpha D  = 2e10^{-5}  (m^3) 
\end{cases}
\end{equation}

\subsection{Optimization Approaches} 
\label{Optimization Modification}
In this paper, we used six optimization methods for the described problem. Three of them are taken from the numerical methods and the other three are from metaheuristic methods. A brief description is given for each of the methods.
\subsubsection{Nelder-Mead}
One of the most popular numerical methods is Nelder-Mead which is based on the concept of the simplex method. To solve a problem with $n$ dimensions, Nelder-Mead keeps $n+1$ test points to approximate a local optimum of a problem. Noting that the maximum number of evaluations in each iteration is not more than two. The main part of the algorithm is extrapolating the value of the objective function in each test point to find a new test point. Then the new test point is replaced with the older test points. This procedure continues to achieve the solution. A simple way of extrapolating is to replace the worst point in terms of objective value with a reflected point from the center of the remaining n points. Two approaches are proceeding when extrapolating is done. if the reflected point has the better answer, then the expanded point is computed, however, if the reflected point is not the better answer, then then the contracted point is computed \cite{avriel2003nonlinear,lagarias1998convergence}.
\begin{equation}
    \begin{aligned}\label{eq:reflected}
   &  X_{reflected} =X_0+\alpha (X_0-X_{n+1} )\    , \ \  \quad  \quad \quad \alpha >o  \\
   &  X_{expanded}=X_0+\gamma(X_r-X_0) \   ,\quad \quad \quad \quad \gamma >1  \\
   &  X_{contracted}=X_0+ \rho (X_{n+1} -X_0) \   , \quad 0< \rho <= 0.5
\end{aligned}
\end{equation}
\subsubsection{Interior-Point Method}
Another algorithm selected in this study is the active-set method that is one of the iterative optimization methods. Active-set methods have two iterative sections. The first is including a feasible region and feasible initial guess. The latter is the phase that the objective is optimized for nonempty feasible region \cite{vaccari2019sequential}. The sequence of feasible iterates $X_{k}$ can be computed according to the equation below.

\begin{equation} \label{active-set}
X_{k+1} = X_{k} + \alpha_k p_k~~~~~where: F_{X_{k+1}}>F_{X_k}
\end{equation}

\noindent where $\alpha_k $ is positive step length, $p_k$ is nonzero search direction and $F_{X_{k+1}}$ is the cost function value \cite{wong2011active}.

\subsubsection{Sequential Quadratic Programming}

To solve a constrained nonlinear optimization problem, an iterative method called Sequential Quadratic Programming (SQP) can be used in which the objective function together with the constraints are smooth nonlinear function which means can be at least twice continuously differentiable. The main idea is to optimize the quadratic model of the objective function while considering the linearized constraints. Generally, this algorithm is the enhanced version of Newton’s method. To begin with, the Lagrangian of the problem and the hessian of the Lagrangian must calculate, then identify the jacobian matrix of the constraints. After that, the given problem is a nonlinear problem that solves within the algorithm \cite{boggs1995sequential}.






\subsubsection{Multi-Verse Optimizer}

An interesting algorithm inspired by cosmology and consists of a white hole, black hole, and wormhole called Multi-Verse Optimizer (MVO). The concept of the algorithm is transferring objects from the white hole (the universe with a higher inflation rate has more probability to have a white hole) to the black holes. The inflation rate refers to the solution of the fitness function.  The Roulette wheel was used to designate a white and black whole to the universes. Because of transferring objects from higher to lower inflation rate universes, the average inflation rate improves as the algorithm runs. It is noted that a wormhole that transfers objects from the best universe to another universe is used for enhancing the exploitation part \cite{mirjalili2016multi}.

\begin{equation}
    \begin{cases}
X_m^n =X_n+TDR(Ur_n-Lr_n)&Ran3<0.5\\
\qquad \qquad \qquad \qquad \qquad \qquad \qquad \qquad Ran2<WEP\\ 
X_n-TDR (Ur_n-Lr_n) Ran4 + Lr_n) & Ran3 > 0.5
\\

X_m^n\qquad \qquad \qquad \qquad \qquad \qquad \qquad \qquad Ran2>WEP \\
    \end{cases}
\end{equation}
Where $X_n$ is the nth parameter of the Best Universe so far, $TDR$ is the traveling distance rate, $WEP$ is the wormhole existence probability, and $Lr_n$, $Ur_n$ are the lower and upper bounds of nth variable, $x_m^n$ is the nth parameter of the mth Universe and $ran2$, $ran3$ and $ran4$ are random numbers between 0 and 1. 

%




\subsubsection{Modified combination of Genetic, Surrogate and fminsearch algorithms}
The genetic algorithm (GA) is one of the most famous evolutionary algorithms that has been used largely for wave energy applications \cite{garcia2020hull}. It is a metaheuristic algorithm that is inspired by natural selection. This algorithm can solve global optimization problems and it is particularly effective when the gradient of the objective function is not available. Furthermore, it can be used in a parallel computation reducing the computational time required. The simulation of each individual of the population is independent and could be run in parallel. The population at each generation is improved using three main operators that are the selection, cross-over, and mutation. The Stochastic Universal Sampling has been used for the selection operator while the cross-over and the mutation operators have been chosen respectively as the uniform and the real-coded mutation. An open-source MATLAB code has been used as a starting code for this purpose\cite{Blasco2021}. Then, several improvements have been made to this algorithm as explained in \cite{faraggiana2018design,faraggiana2020genetic}. This code has been modified to avoid repetitions and to enable new evaluations in the optimal region found. It is also improved to allow the tuning of the mutation, cross-over probability and the population as described in \cite{faraggiana2020genetic}. Then, an elitism factor has been introduced in the genetic algorithm to insert new individuals in the population from a surrogate model which increases the efficiency of the algorithm. The huge advantage of the surrogate model is that enables the user to estimate the results at a very low computational time compared to the real simulation. The Kriging surrogate model is chosen because it is suitable for highly nonlinear responses. It belongs to the family of linear least square algorithms and the estimation of the response of a point x is based on a linear combination of the results of the simulations and it is expressed as
\begin{equation}
    f(x) = \sum_{i=1}^{N} \lambda_i (x)f(x_i) \\
\end{equation}
where $\lambda_i$ are the weights obtained from a regression analysis. Then, a Gaussian process is built through the residuals. The open-source Oodace toolbox has been used for the surrogate model \cite{couckuyt2014oodace} while a parallel computation of the GA+surrogate has been implemented using the HPC HACTAR of Politecnico of Torino \cite{Torino2021}. Finally, the MATLAB "fminsearch" function has been added after the GA+surrogate optimization to allow the final convergence of the optimization. The function "fminsearch" uses the Nelder-Mead simplex method that is very efficient to find the local optimum. So, this final algorithm is suitable to refine the optimum found after that the design space has been explored by the previous global optimization algorithm.\\

\section{Feasibility Study and Sensitivity Analysis} 
In order to obtain the applicable range of the decision variables for the optimization problem, a few relevant studies were used, in particular two studies that had the identical point absorber as this study. In \cite{yu2018numerical} efficiency of an HPTO for wave energy converters and the trade-off between generated power and oscillation between different power smoothing techniques was assessed. In \cite{quartier2018numerical} a numerical model of a point absorber using HPTO with three different control strategies was investigated. These two studies were the main reference for the selected range of the 4 HPTO parameters. Also, in \cite{casey2013modeling} another point absorber was studied which was considerably smaller than our model, but had a higher value for the High-Pressure Gas Accumulator volume, which was taken into consideration for finalizing the ranges. These ranges can be seen in Table \ref{table:SenAnMetricsPre}. 

In the next step, we did a sensitivity analysis using the obtained ranges from the literature review, in which we ran the simulation with different values of each decision variable. We did a total of 266 runs in this section. A summary of these runs are presented in Table \ref{table:SenAnMetricsPre}. In a number of cases, the values of most outputs were not applicable in practice, especially the values for piston pressures, the PTO force, absorbed power, and displacement of the device. There were a total of 41 cases with these described properties. After further investigations, a pattern was found in these cases, the values of Piston Area and HPA Volume in these cases had something in common, the volume of HPA Volume was lower than a certain threshold. The minimum acceptable value of HPA Volume was found for different Piston Areas and the acceptable values were found. 

\begin{table*}[!htb]
\centering
\caption{ Metrics of Sensitivity Analysis Runs Before Data Reduction. (consisting of 266 cases) }
\label{table:SenAnMetricsPre}
 \scalebox{0.7}{
\begin{tabular}{p{16mm}p{9mm}p{7mm}p{7mm}p{14mm}p{12mm}p{14mm}p{13mm}p{13mm}p{11mm}p{9mm}p{15mm}p{15mm}p{15mm}}

\hlineB{5}

\textbf{Parameters}   & \textbf{Piston Area (m$^2$)} & \textbf{HPA Volume (m$^3$)}  & \textbf{LPA Volume (m$^3$)}  & \textbf{LPA Pre-Charge Pressure (MPa)} & \textbf{Max Piston Pressure (MPa)} & \textbf{Min Piston Pressure (MPa)} & {\textbf{Max PTO Force (MN})} & \textbf{Mean Absorbed Power (kW)} & {\textbf{Mean Generated Power (kW)}} & {\textbf{Mean Electrical Power (kW)}} & {\textbf{Max Float Displacement (m)}} & {\textbf{Max Spar Displacement (m)}}\\ \hlineB{2}

\textbf{Max}	 & 0.18	& 10	& 8	& 9.6 &	50428.55	& 17.90	& 204883.44 &	39323.99 &	343.10	& 275.75	& 435.55 &	780.67
           \\ \hline %
\textbf{Min}	 & 0.045	& 0.5	& 0.5	& 3.5	& 12.71	& $-370852.02$	& 0.39	& 84.62 &	68.73 &	59.62 &	3.23 &	1.25
           \\ \hline %
\textbf{Average}	& 0.103 &	5.03	& 4.80 &	6.27 &	655.80 & $-1485.99$ &	1209.25 &	452.55 &	166.69 &	138.57 &	7.70 &	9.53
           \\ \hline %
\textbf{Standard Deviation}	& 0.032 &	2.62 &	1.87 &	1.49 &	4374.61	& 22774.38 &	13175.43 &	2438.14	& 41.39 &	32.65 &	29.03 &	51.00 
                    \\ \hline
 \hlineB{5}         
\end{tabular}
}
\end{table*}

After removing these 41 cases from our database, the same statistics were gathered for the remaining 225 cases in Table \ref{table:SenAnMetricsPost} .

\begin{table*}[!htb]
\centering
\caption{ Metrics of Sensitivity Analysis Runs After Data Reduction.  (consisting of 225 cases) }
\label{table:SenAnMetricsPost}
 \scalebox{0.7}{
\begin{tabular}{p{16mm}p{9mm}p{7mm}p{7mm}p{14mm}p{12mm}p{14mm}p{13mm}p{13mm}p{11mm}p{9mm}p{15mm}p{15mm}p{15mm}}

\hlineB{5}

\textbf{Parameters}   & \textbf{Piston Area (m$^2$)} & \textbf{HPA Volume (m$^3$)}  & \textbf{LPA Volume (m$^3$)}  & \textbf{LPA Pre-Charge Pressure (MPa)} & \textbf{Max Piston Pressure (MPa)} & \textbf{Min Piston Pressure (MPa)} & {\textbf{Max PTO Force (MN})} & \textbf{Mean Absorbed Power (kW)} & {\textbf{Mean Generated Power (kW)}} & {\textbf{Mean Electrical Power (kW)}} & {\textbf{Max Float Displacement (m)}} & {\textbf{Max Spar Displacement (m)}}\\ \hlineB{2}

\textbf{Max}	 & 0.18 &	10 &	8	& 9.6 &	201.51 &	15.76 &	19.92 &	385.53 &	229.19 &	188.11 &	8.28 &	8.64
\\ \hline
\textbf{Min} &	0.045 &	1.5	& 0.5 &	3.5 &	12.71 &	0.37 &	0.39 &	84.62 &	68.73 &	59.62 &	3.29 &	1.25
\\ \hline
\textbf{Average}	& 0.103 &	5.79 &	4.74 &	6.29 &	53.63 &	5.97 &	5.29 &	246.78 &	161.90 &	135.02 &	4.04 &	3.61
\\ \hline
\textbf{Standard Deviation}	& 0.033 &	2.08 &	1.92 &	1.54 &	51.60 &	2.87 &	5.50 &	73.54 &	37.79 &	29.98 &	1.36 &	1.84  
                    \\ \hline 
 \hlineB{5}      
\end{tabular}
}
\end{table*}
By comparing Tables \ref{table:SenAnMetricsPre} and \ref{table:SenAnMetricsPost}, it can be seen that there is not much change in the column corresponding with the four decision variables, namely piston area, HP- and LP accumulators volumes, and LPA pre-charge pressure. But for previously problematic outputs, there is a significant change. The value for the maximum of max piston pressure has been decreased from 50428 MPa to 201 MPa and the minimum of min piston pressure has been increased from -370852 MPa to 0.37 MPa. Also, the values for these two parameters' average and standard deviation are smaller and more reasonable. Data regarding the max PTO force, average absorbed power and device’s displacement has been significantly improved mainly in case of their standard deviation and some in case of their average value. Also, the maximum value for most variables in the table has been decreased by at least 2 orders of magnitude. Overall we can see that the data reduction was necessary, and it is considered a success.

\begin{figure}[H]
    \centering
    \captionsetup{justification=centering}
    \includegraphics[width=1\linewidth]{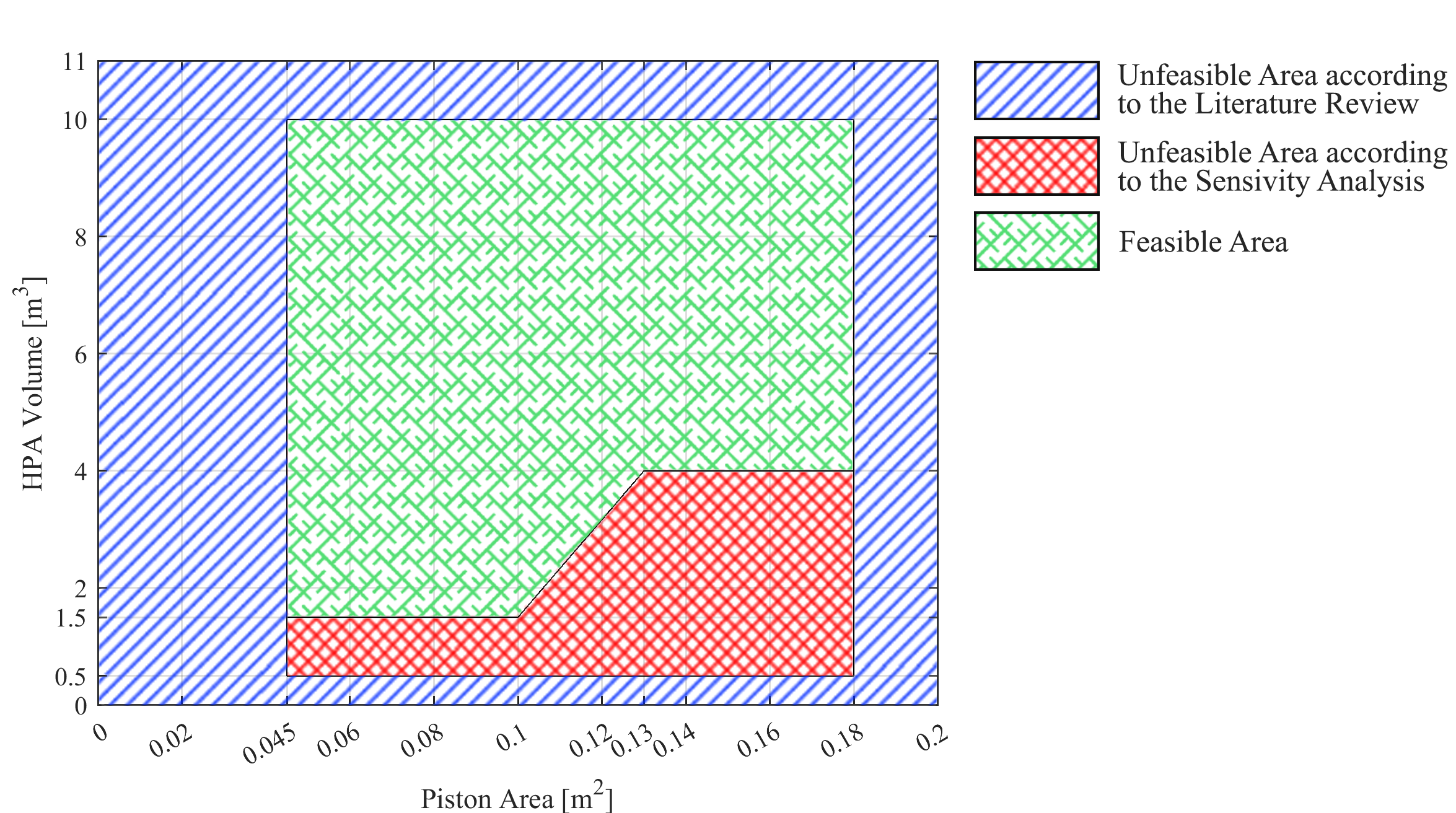}
    \caption{Feasibility of Search Space of Piston Area and HPA Volume of the HPTO}
       \label{SA_FeasibleArea}
\end{figure}

After investigating the effect of data reduction, and making sure that it was helpful, we finalized the penalty area of our search space. This has been illustrated in Fig. \ref{SA_FeasibleArea}, in which the green section represents the feasible area in this study.

\begin{figure}[H]
    \centering
    \captionsetup{justification=centering}
    \includegraphics[width=1\linewidth]{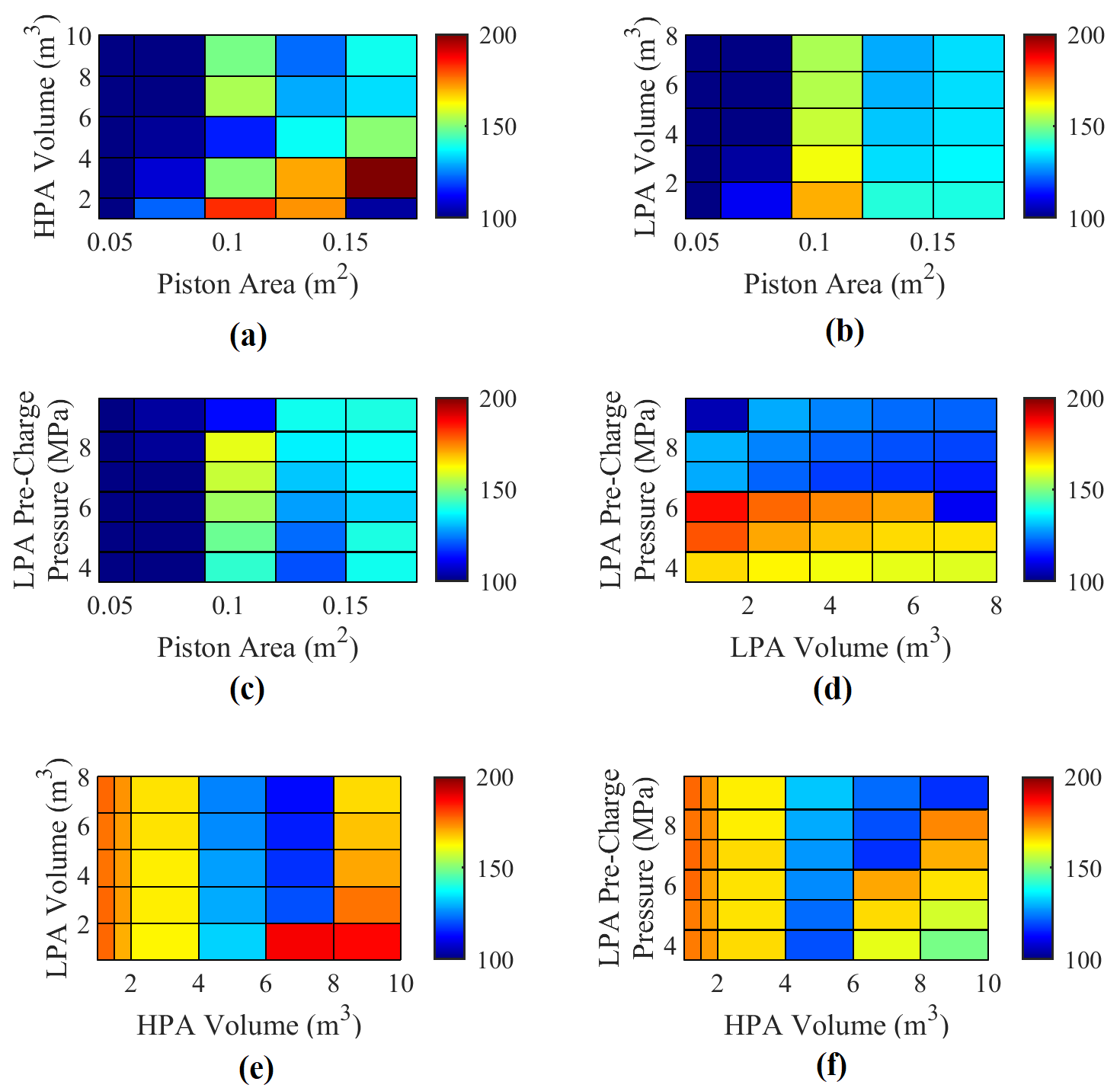}
    \caption{Sensitivity Analysis Plots for the Effects of Changes in the 4 Decision Variables Values on the Output Power}
       \label{6SA_Plots}
\end{figure}

In the next step, we investigated the rough influence of HPTO parameters values in groups of 2 on the mean output power. This resulted in 6 plots which are presented in Fig. \ref{6SA_Plots}.

As it is apparent in the first 3 plots ((a), (b) and (c)) in Fig. \ref{6SA_Plots} that include piston area, a value below 0.1 m$^2$ for this parameter results in a significant drop in the power output, this is consistent with \cite{gaspar2018design}. Also, according to Fig. \ref{6SA_Plots}-a, the best results happen when the piston area is relatively large and the HPA volume is relatively small, and if both reach the extremum value in the feasible area the power output drops. Additionally, based on Fig. \ref{6SA_Plots}-b, for the same values for the low-pressure gas accumulator volume, the increase in the piston area generally improves the power output, except in the cases with LPA volume below 2 m$^3$.

According to Fig. \ref{6SA_Plots}-d, when the LPA pre-charge pressure is below 6.5 MPa, the objective function is generally better, and it has especially good values when LPA volume is low and LPA pre-charge pressure has a moderate value with respect to our feasible range. As can be seen in Fig. \ref{6SA_Plots}-e, if the HPA volume is constant, the LPA volume does not affect the absorbed energy notably, except when the former is large, and the latter has a small value, which results in exceptionally good answers. Finally, according to Fig. \ref{6SA_Plots}-f, for the low values of high-pressure gas accumulator volume, LPA pre-charge pressure is not as influential as HPA volume in the power output.

The fluctuating nature of the ocean waves is one of the main challenges in utilizing wave energy devices. Consequently, WEC components have to be designed to handle loads bigger than the average load, which subsequently decreases mechanical and volumetric losses and increases the Levelised Cost Of Energy (LCOE) of the WEC. WEC’s maximum power output can reach one order of magnitude greater than the mean power output \cite{yu2018numerical}.
Another disadvantage of these fluctuations in an energy type is that it makes that energy an unfitting option for sensitive equipment that needs a stable voltage. For these reasons, the peak-to-average power ratio should be decreased while maintaining a high average of absorbed energy. This can be done by using accumulators and advanced control methods. However, it should be noted that the cost of the PTO is not directly proportional to the maximum absorbed energy, especially for hydraulic systems similar to those investigated here \cite{yu2018numerical}.
In order to assess fluctuations of the HPTO power output, we determine the power fluctuation ratio \cite{yu2018numerical}:

\begin{equation}
R_{PF} = \frac{\Delta P_E}{P_{avg}} = \frac{P_{max} - P_{min}}{P_{avg}}
\end{equation}

\noindent where the $P_{max}$ and $P_{min}$ values are calculated using 99.9 and 0.1 percentiles of power peaks during the simulation time \cite{yu2018numerical}.

\begin{figure}[H]
\centering
\captionsetup{justification=centering}
\includegraphics[width=\linewidth]{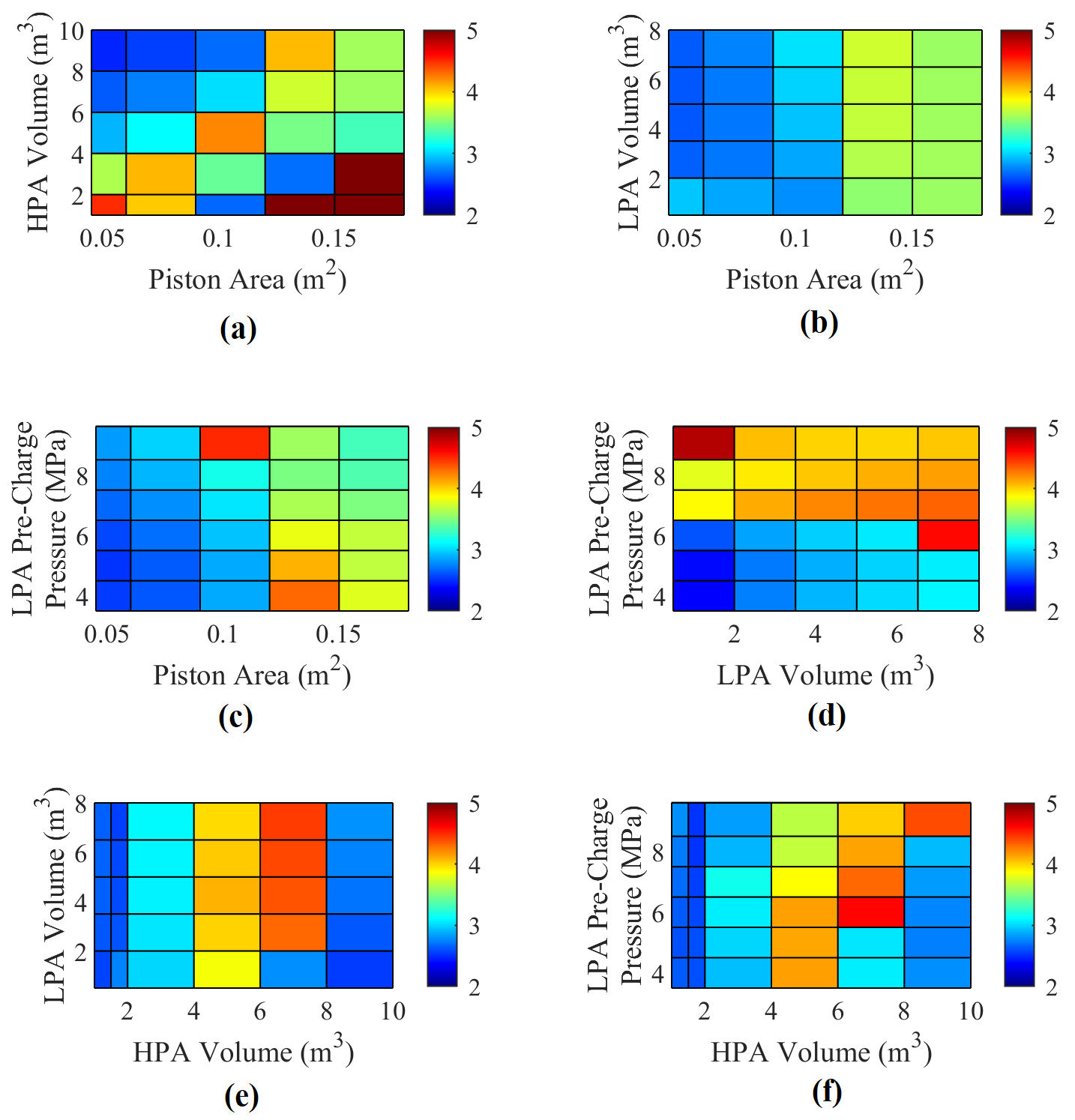}
\caption{Sensitivity analysis plots for the effects of changes in the 4 decision variables values on the power fluctuation ratio ($R_{PF}$)}
\label{6SA_RPF}
\end{figure}

To investigate the effects of changing the HPTO parameters on the power fluctuation, a similar sensitivity analysis to Fig. \ref{6SA_Plots} are done in Fig. \ref{6SA_RPF} but this time for the $R_{PF}$.

According to Fig. \ref{6SA_RPF}-(a), low volumes of HPA intensify the $R_{PF}$ drastically, especially when its value is below 4 m$^3$.  If you take a closer look at the plots that investigate LPA volume (plots (b), (d), and (e)), you can see that the LPA volume does not seem to influence the $R_{PF}$ significantly, except in a few cases, for example, when LPA is very large or very small, and LPA pre-charge pressure is close to its maximum feasible value. Fig. \ref{6SA_RPF}-(b) and (c) show that low values for the piston area can actually be better for the power fluctuation ratio.
Fig. \ref{6SA_RPF}-(c) and (f) also show that LPA pre-charge pressure increase can have opposite effects on the $R_{PF}$ depending on the piston area value. Based on Fig. \ref{6SA_RPF}-(d), the low values for the LPA pre-charge pressure result in lower $R_{PF}$s, which is more advantageous. And as noted before, small volume and large pre-charge pressure for the LPA can be undesirable. By analyzing the last two plots of Fig. \ref{6SA_RPF} (plots (e) and (f)), one can see that moderate amounts for the HPA volume result in large $R_{PF}$s, which is not preferable.

\section{Optimization Convergence}
\label{Optimization Convergence}
\subsection{Numerical Methods}
Three numerical optimization algorithms, namely Nelder-Mead, SQP, and Active-set, are used to solve our optimization problem. The properties of these 3 runs are presented in table \ref{table:NumConvDet}. According to this table, Nelder-Mead had the most number of iterations before reaching the termination criteria, which is almost 5 times bigger than the iterations Active-set needed to reach the said criteria. Since we had an unfeasible area for our inputs, some of the iterations in these 3 runs are not acceptable. Active-set had the most number of unfeasible iterations despite having the least number of total iterations, with a 9.8 \% of evaluations being unfeasible. Next is SQP which had a 1.5 \% of unfeasible iterations, and lastly Nelder-Mead had 0 unfeasible iterations which is very good.

\begin{figure}[H]
    \centering
    \captionsetup{justification=centering}
    \includegraphics[width=0.8\linewidth]{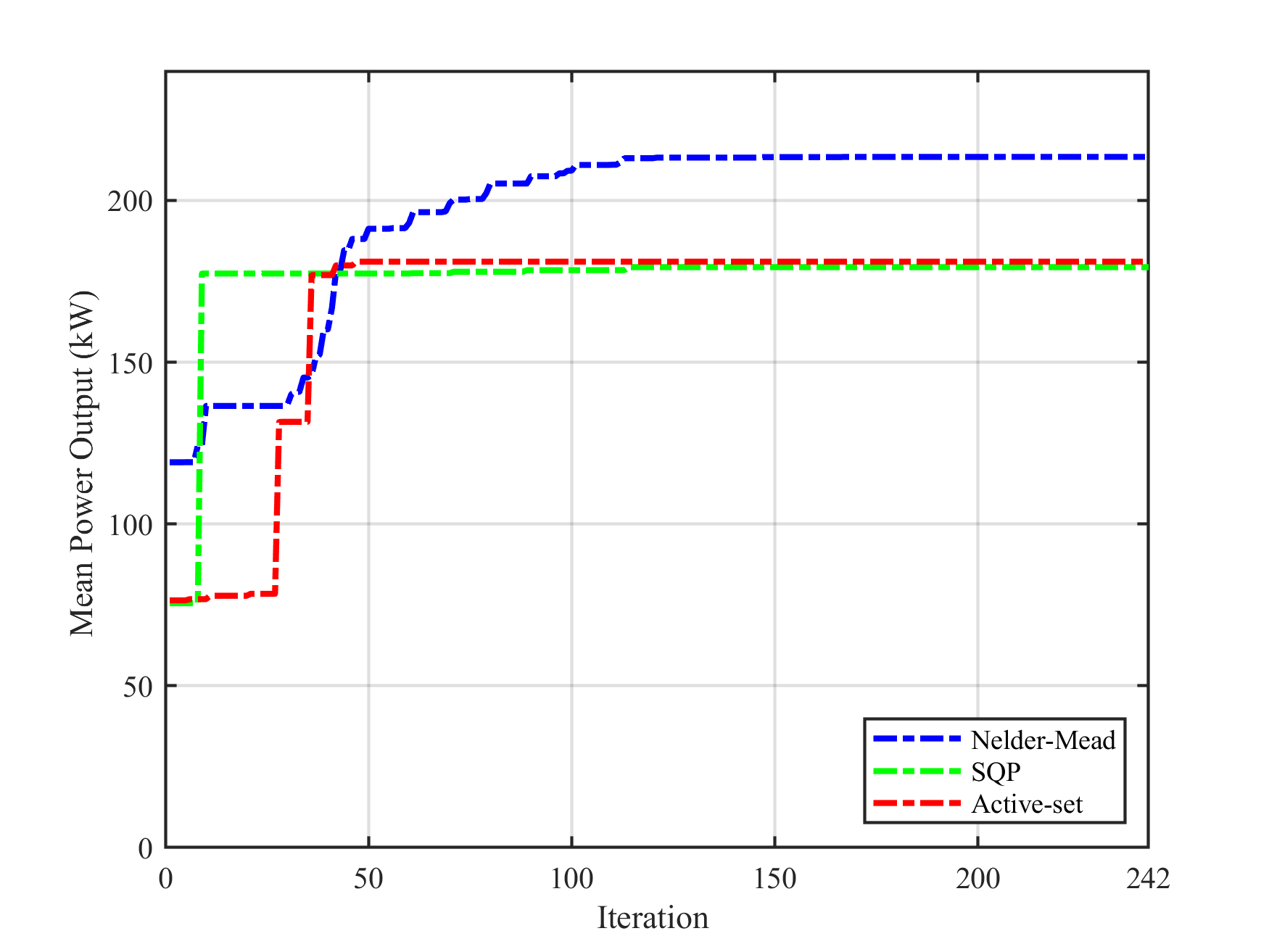}
    \caption{Convergence curve of the mean output power for the 3 numerical optimization algorithms (Nelder-Mead, SQP, and Active-set)}
       \label{CC_Power_NUM}
\end{figure}

The optimization convergence results are presented in table \ref{table:OptRes} and Fig. \ref{CC_Power_NUM}. The HPTO parameters and power output does not show a very good convergence overall. The only exception is the LPA pre-charge pressure, which range of optimal parameters is 2.6 \% of the total feasible design range for that parameter.  Also, they have converged next to the lowest value for the feasible range. The LPA volume has the least convergence, its range of optimal parameters is 99 \% of the total feasible design range for LPA volume. Finally, If Nelder-Mead is excluded, the other two algorithms have pretty good convergence in piston area and the HPA volume, the range of optimal values according to these algorithms are 3 \% and 0.1 \% of the total possible design range, respectively. 

\begin{table*}[!htb]
\centering
\caption{ Numerical Runs Details.}
\label{table:NumConvDet}
 \scalebox{0.8}{
\begin{tabular}{p{45mm}p{35mm}p{35mm}p{20mm}}

\hlineB{5}

\textbf{Numerical Algorithm Name}   & \textbf{Total Iterations} & \textbf{Unfeasible Iterations}   \\ \hlineB{2}

\textbf{Nelder-Mead}  & 242 &	0  \\
\hline %
\textbf{SQP}  &	195 &	3  \\
\hline 
\textbf{Active-set}  &	51 &	5 			\\
\hline 
\hlineB{5}   
\end{tabular}
}
\end{table*}

\subsection{Modified Combined Method}
\begin{figure}[H]
    \centering
    \captionsetup{justification=centering}
    \includegraphics[width=0.8\linewidth]{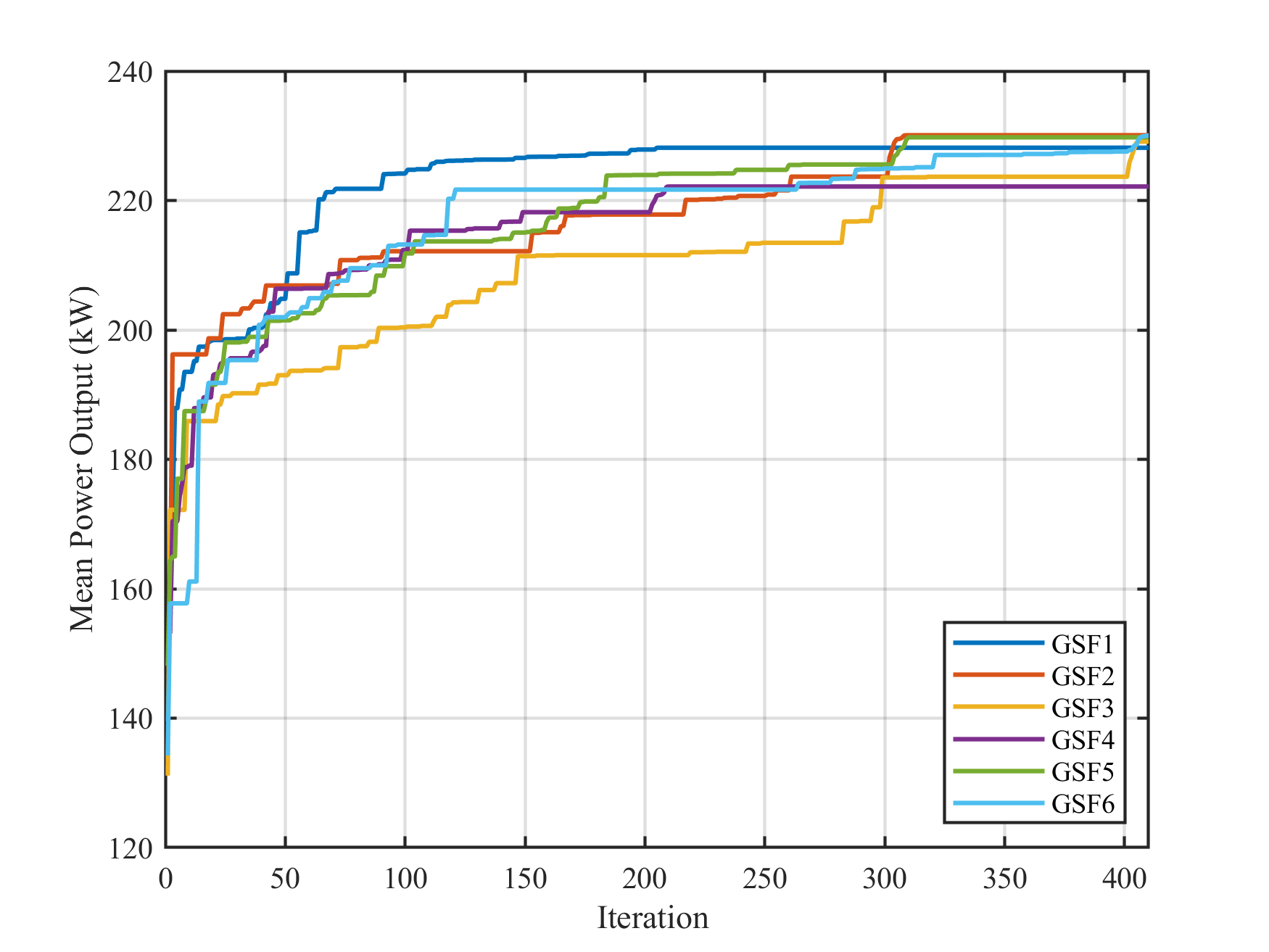}
    \caption{Convergence curve of the mean output power for the 6 GSF (GA+surrogate+fminsearch) optimization algorithms. A convergence step of 10 simulations is used for visualization purposes.}
       \label{CC_Power_GSF}
\end{figure}

The optimization convergence has been checked for six optimization runs and their optimization settings are defined in Table \ref{table:OptConvSet}. The population and the surrogate elitism factor have been investigated with three and two different values respectively. The number of generations of the GA and the number of maximum evaluations of the fminsearch function have been instead fixed to 50 and 100 respectively. The total number of simulations is then determined by the sum of the product between the population and the number of generations and the number of evaluations of fminsearch. The optimisation convergence results are shown in Table \ref{table:OptRes} and in Fig. \ref{CC_Power_GSF}. The design parameters and the electrical power demonstrate a good convergence from these results. In fact, the design parameters of each optimization run have converged to similar design parameters. The HPA and LPA volumes design parameters have converged really well, especially the LPA volume. The range of the optimal parameters of these variables are respectively 2.7\% and 0.3\% of the full design space range for the HPA and LPA volumes. They have converged also next to the highest and lowest values of the range considered respectively. The least converged design parameters are the piston area and the LPA pre-charge pressure. In fact, the range of these optimal parameters is around 12\% and 13\% respectively. However, this is mostly due to GSF1 and GSF4 that are the least converged optimization runs, probably due to their lowest number of total simulations. The converged range of these variables decreases to only around 3\% if these two optimizations runs are not accounted for. Table \ref{table:OptRes} shows also the minimum and maximum piston pressure, the maximum PTO force, the power and the maximum displacement of the float and the spar. The maximum and the minimum piston pressure are respectively the most and the least converged results among them. The optimal absorbed, generated and electrical powers give an optimal electrical efficiency of around 72-75\%. Fig. \ref{CC_Power_GSF} shows the convergence plot of the electrical power for the six optimization runs. Each optimization run has a different number of simulations with GSF3 and GSF6 that have the largest ones. The best results are given from GSF2 and GSF6 in which the optimal electrical power converges around 230 kW.

\begin{table*}[!htb]
\centering
\caption{ Convergence set-up for six optimisation runs.}
\label{table:OptConvSet}
 \scalebox{0.8}{
\begin{tabular}{p{16mm}p{16mm}p{25mm}p{35mm}p{20mm}p{30mm}}

\hlineB{5}

\textbf{Run}   & \textbf{npop} & \textbf{surr. elit. factor}  & \textbf{ngen (GA+surrogate)}  & \textbf{nfminsearch} & \textbf{Total simulations}\\ \hlineB{2}

\textbf{GSF1}	 & 40	& 0.1	& 50	& 100 &	2100
           \\ \hline %
\textbf{GSF2}	 & 60	& 0.1	& 50	& 100	& 3100	
           \\ \hline %
\textbf{GSF3}	& 80 &	0.1	& 50 &	100 &	4100 
           \\ \hline %
\textbf{GSF4}	& 40 &	0.2 &	50 &	100 &	2100	
                    \\ \hline
\textbf{GSF5}	& 60 &	0.2 &	50 &	100 &	3100	
                    \\ \hline
\textbf{GSF6}	& 80 &	0.2 &	50 &	100 &	4100	
                    \\ \hline
\hlineB{5}   
\end{tabular}
}
\end{table*}

\subsection{Multi-Verse Optimizer (MVO)}
\begin{figure}[H]
    \centering
    \captionsetup{justification=centering}
    \includegraphics[width=0.8\linewidth]{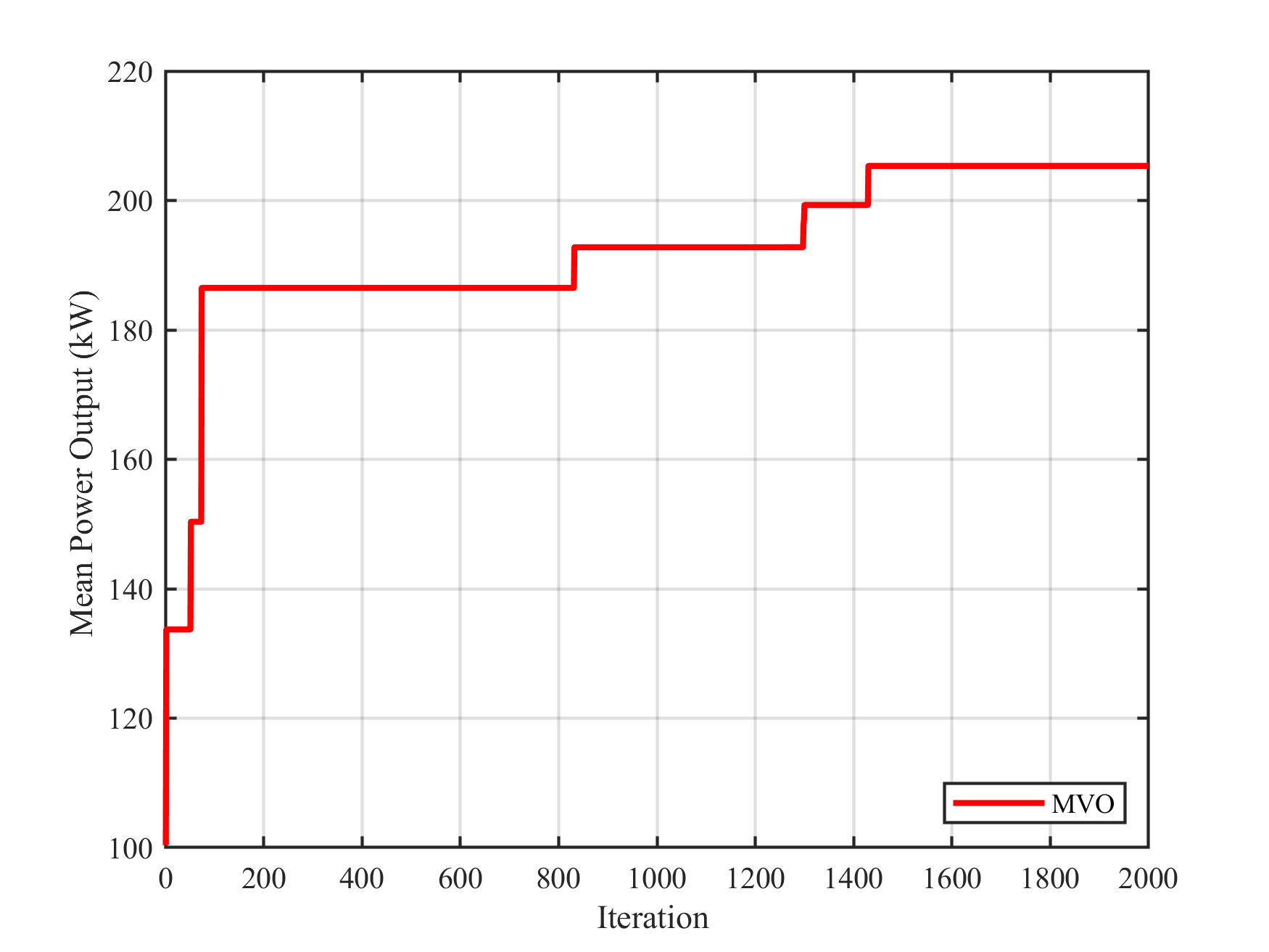}
    \caption{Convergence curve of the mean output power for the MVO (Multi-Verse Optimization) algorithm}
       \label{CC_Power_MVO}
\end{figure}

Lastly, the Multi-Verse Optimization algorithm is used in order to find the optimum parameters of an HPTO unit. The results are shown in Table \ref{table:OptRes} and Fig. \ref{CC_Power_MVO}. The number total of iterations was 2000 and the number of search agents was 10, so in this run, a total number of 20000 evaluations was performed. This algorithm by far had the most number of iterations but did not result in the best possible solution.

\section{Comparative Assessment} 

\label{Comparative Assessment}
In this section, the performance of the 10 optimization algorithms in this paper was compared. This was done by plotting the chronological variation of the objective function and the 4 variables of the 10 algorithms in a figure. Since the Multi-Verse Optimization algorithms had the most iterations between our algorithms, 2000 iterations, and this was about 4 times more iterations than the GSF3 and GSF6 which had the next more iterations, we decided to cut the curves corresponding to the MVO ‘s performance and only show it until the 410th iteration (which is the same iterations as GSF3 and GSF6). As for the other algorithms, we extended their curves horizontally to meet the 410th iteration mark. This was done only for illustration and comparison purposes.

Here we present the best parameters obtained by each optimization algorithm, and the non-optimized case, and other performance parameters in Table \ref{table:OptRes}

\begin{table*}[!htb]
\centering
\caption{ Properties of the non-optimized case and the 10 optimized cases using 10 different optimization algorithms}
\label{table:OptRes}
 \scalebox{0.9}{
\begin{tabular}{p{16mm}p{9mm}p{7mm}p{7mm}p{14mm}p{13mm}p{12mm}p{10mm}p{12mm}p{9mm}p{9mm}p{13mm}p{15mm}p{14mm}p{10mm}p{10mm}}

\hlineB{5}

\textbf{Run}   & \textbf{Number of Iterations} & \textbf{Piston Area (m$^2$)} & \textbf{HPA Volume (m$^3$)}  & \textbf{LPA Volume (m$^3$)}  & \textbf{LPA Pre-Charge Pressure (MPa)} & \textbf{Mean Absorbed Power (kW)} & \textbf{Mean Generated Power (kW)} & \textbf{Mean Electrical Power (kW)} & \textbf{Power Fluctuation Ratio} \\ \hlineB{2}

\textbf{Non-Optimized Case}	& - &	0.038 &	8.50 &	6.00 &	9.60 &	74 &	65 &	57 & 2.63
\\ \hline
\textbf{Nelder-Mead}	& 242 &	0.146 &	9.49 &	0.58 &	3.66 &	320 &	263	& 213 & 2.47
\\ \hline
\textbf{SQP}	& 195 &	0.091 &	1.51 &	4.05 &	3.50 &	285 &	221 &	179 & 2.91
\\ \hline
\textbf{Active-set} &	51 &	0.087 &	1.5 &	8 &	3.50 &	288 &	223 &	181 & 2.84
\\ \hline
\textbf{GSF1} &	210 &	0.132 &	9.97 &	0.51 &	5.29 & 310 &	281 &	228 & 2.31
\\ \hline
\textbf{GSF2}	& 310 &	0.140 &	9.99 &	0.52 &	4.85 &	320 &	284 &	230 & 2.29
\\ \hline
\textbf{GSF3}	& 410 &	0.136 &	9.76 &	0.50 &	5.03 &	314 &	282 &	229 & 2.30
\\ \hline
\textbf{GSF4} &	210 &	0.124 &	9.75 &	0.50 &	5.64 &	297 &	273 &	222 & 2.38
\\ \hline
\textbf{GSF5}	& 310 &	0.137 &	9.91 &	0.50 &	5.06 &	315 &	283 &	230 & 2.30
\\ \hline
\textbf{GSF6} &	410 &	0.140 &	9.96 &	0.52 &	4.85 &	320 &	283 &	230 & 2.30
\\ \hline
\textbf{MVO}	& 2000 &	0.121 &	6.30 &	0.50 &	4.05 &	283 &	251 &	205 & 2.57
\\ \hline
\hlineB{5}   
\end{tabular}
}
\end{table*}

From the sensitivity analysis section, we know that lower values of $A_p$s are not really desirable, which is consistent with all the solutions in Table \ref{table:OptRes}. The two worst performers, i.e., the SQP and Active-set cases, have an $A_p$ below 1 $m^2$. Also, the LPA pre-charge pressure value is not high for the optimized cases, which conforms to Fig. \ref{6SA_Plots}.

\begin{figure}[H]
    \centering
    \captionsetup{justification=centering}
    \includegraphics[width=1\linewidth]{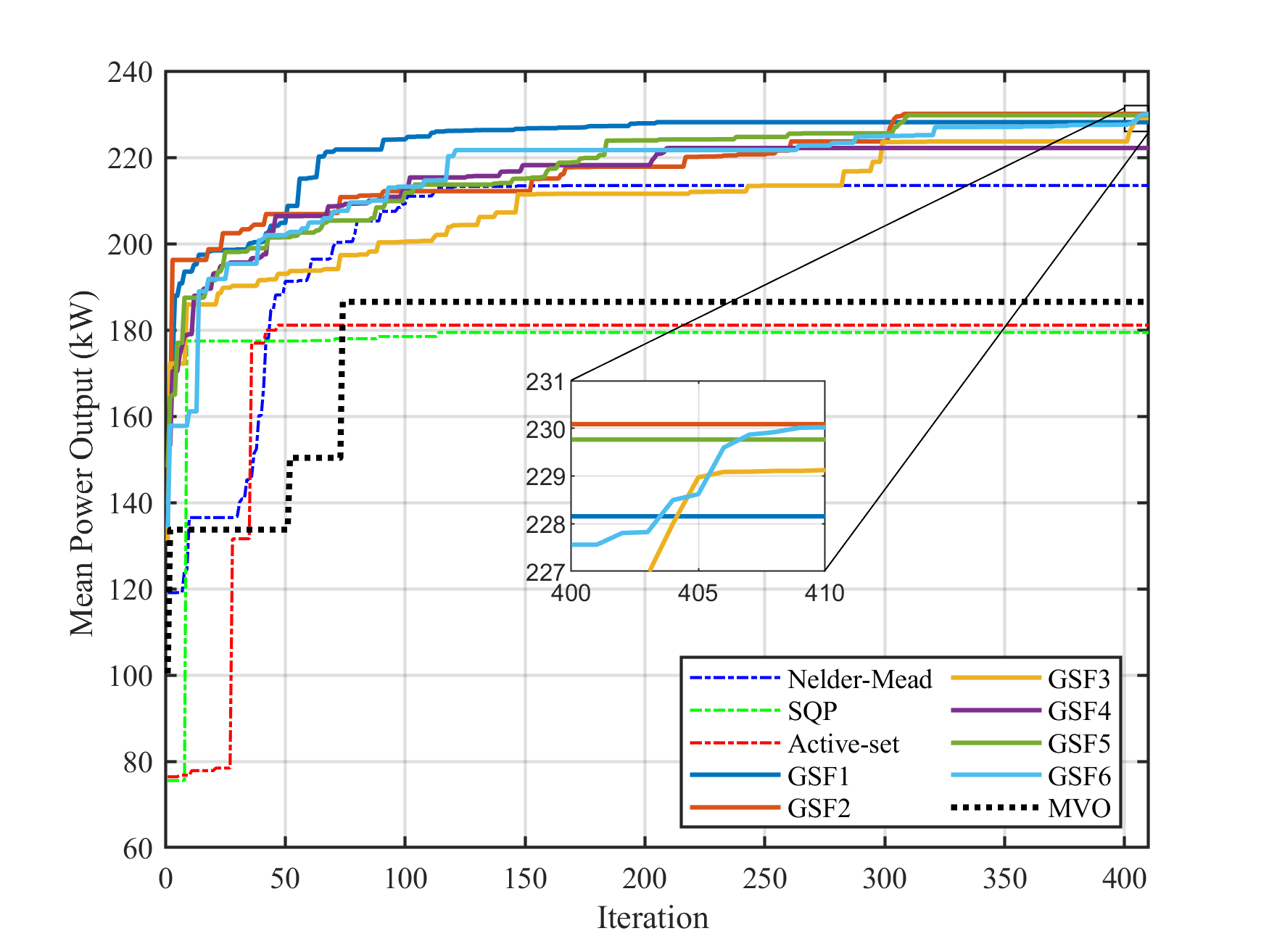}
    \caption{Convergence curve of the mean output power for all the 10 optimization algorithms (6 GSF algorithms, 3 numerical algorithms, and MVO)}
       \label{CC_Power}
\end{figure}

The first plot for this purpose is the mean power output chronological variation for the 10 algorithms which is displayed in Fig. \ref{CC_Power}. 

As can be seen, the 6 GSFs (Modified combination of Genetic, Surrogate and fminsearch algorithms) had the best performance and were able to reach a mean output power of 230 kW, the worst-performing member was the GSF4 with a 222 kW power output. Since this algorithm had the least number of iterations in GSF algorithms, with only 210 iterations, it is understandable that it did not reach the 230 kW mark. Next in line is the Nelder-Mead algorithm with a power output of 213 kW. From this figure, it seems that the HPTO output power approaches the optimum value starting from iteration number 110. Next, we have MVO, with an output power of 205, which is worse than Nelder-Mead’s performance but better than the other 2 numerical algorithms. Also, MVO reached this value after around 75 iterations. Finally, we have SQP and Active-set, which had the worst performance with a similar output power of around 180 kW. Overall GSFs really outshined the other algorithms in this comparison.

\begin{figure}[H]
    \centering
    \captionsetup{justification=centering}
    \includegraphics[width=1\linewidth]{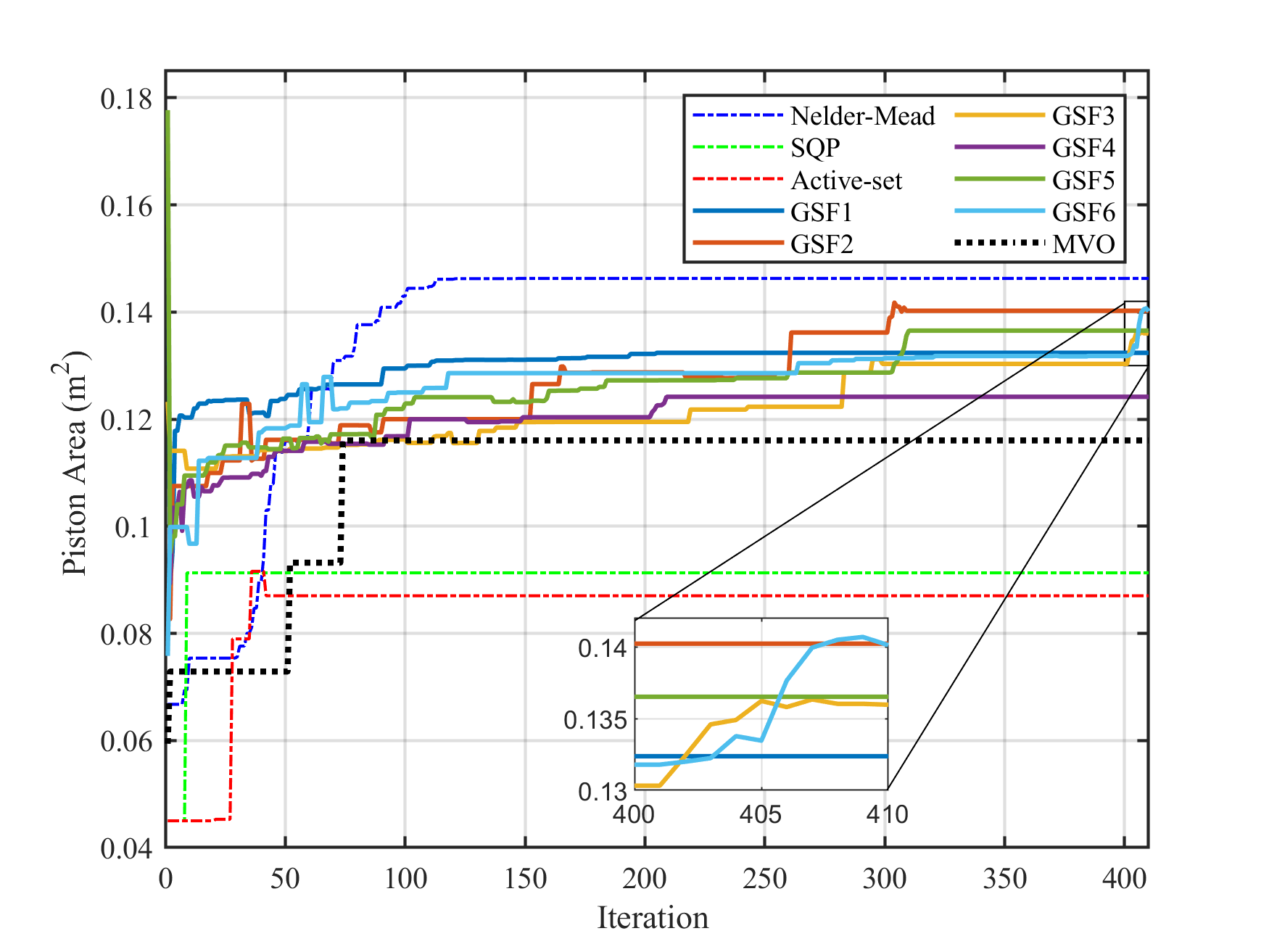}
    \caption{Convergence curve of the first decision variable, the piston area, for all the 10 optimization algorithms (6 GSF algorithms, 3 numerical algorithms, and MVO)}
       \label{CC_Ap}
\end{figure}

Fig. \ref{CC_Ap} shows the chronological variation of the first decision variable, the piston area, for the 10 investigated algorithms. 

The cylinder area did not have the best convergence between the HPTO parameters. All the 6 GSFs converged on a value between 0.124 and 0.14 m$^2$, the Nelder-Mead which is the next best performer had a piston area above this range, 0.146 m$^2$, and the other 3 algorithms reached a piston area below 0.124 m$^2$, the MVO had a relatively similar piston area, but SQP and Active-set had a value around 0.9 m$^2$.

\begin{figure}[H]
    \centering
    \captionsetup{justification=centering}
    \includegraphics[width=1\linewidth]{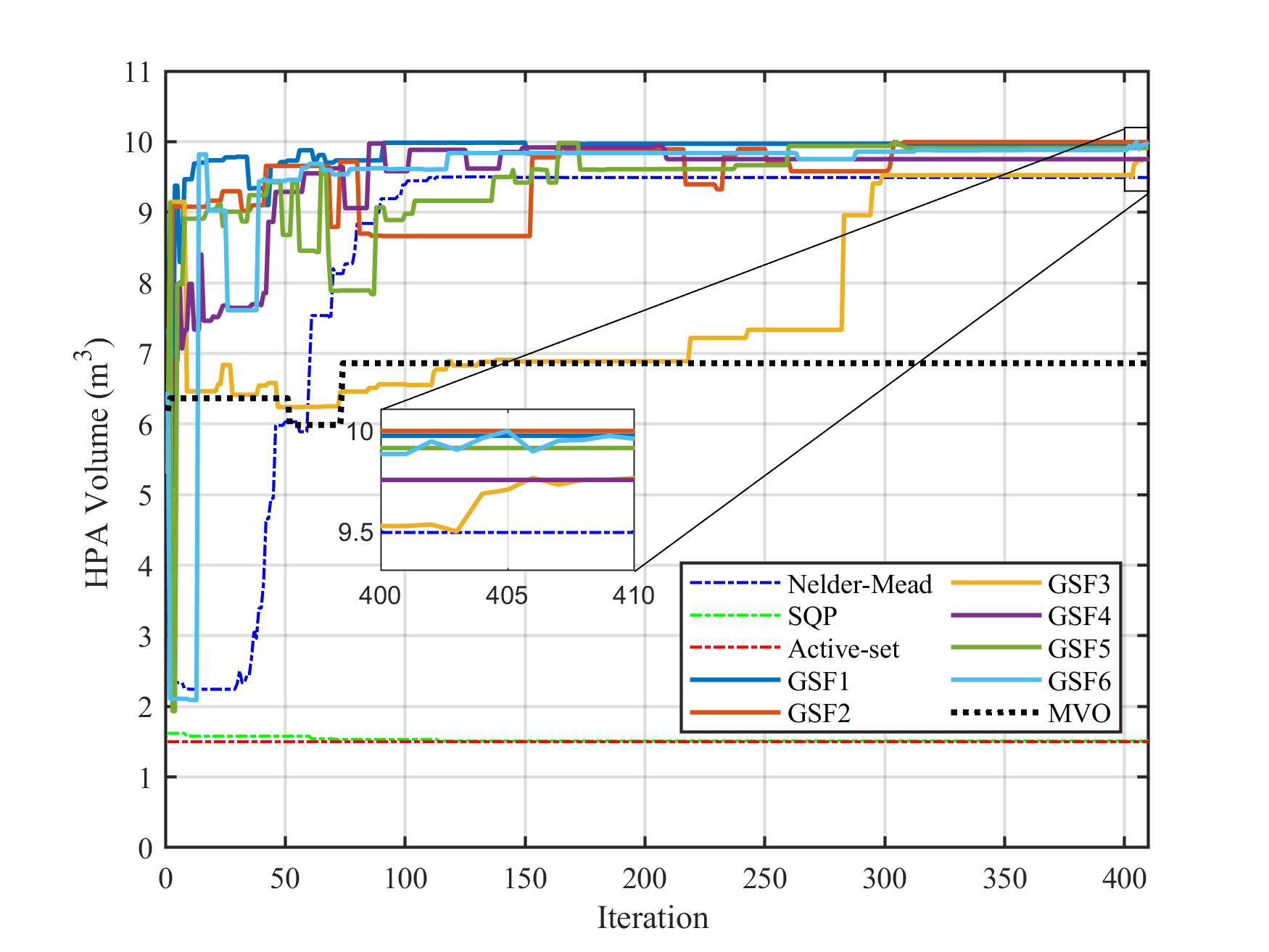}
    \caption{Convergence curve of the second decision variable, the HPA volume, for all the 10 optimization algorithms (6 GSF algorithms, 3 numerical algorithms, and MVO)}
       \label{CC_VH0}
\end{figure}

The next decision variable we plotted the chronological variation of is the high-pressure gas accumulator volume in Fig. \ref{CC_VH0}, which had a very diverse range for the final value of this HPTO parameter.

Similar to the 2 previous plots, the optimum values found by the GSF algorithms were very close, and they ranged from 9.9 to 10 m$^3$, which is near the maximum feasible value for this parameter. Nelder-Mead had an optimum value of 9.5 m$^3$ which was very close. However, the SQP and Active-set algorithms found the optimum HPA volume value to be almost the minimum possible value in the feasible search space with a value of 1.5 m$^3$. And Lastly, the MVO found a 7 m$^3$ volume to be the best for the HPA.

\begin{figure}[H]
    \centering
    \captionsetup{justification=centering}
    \includegraphics[width=1\linewidth]{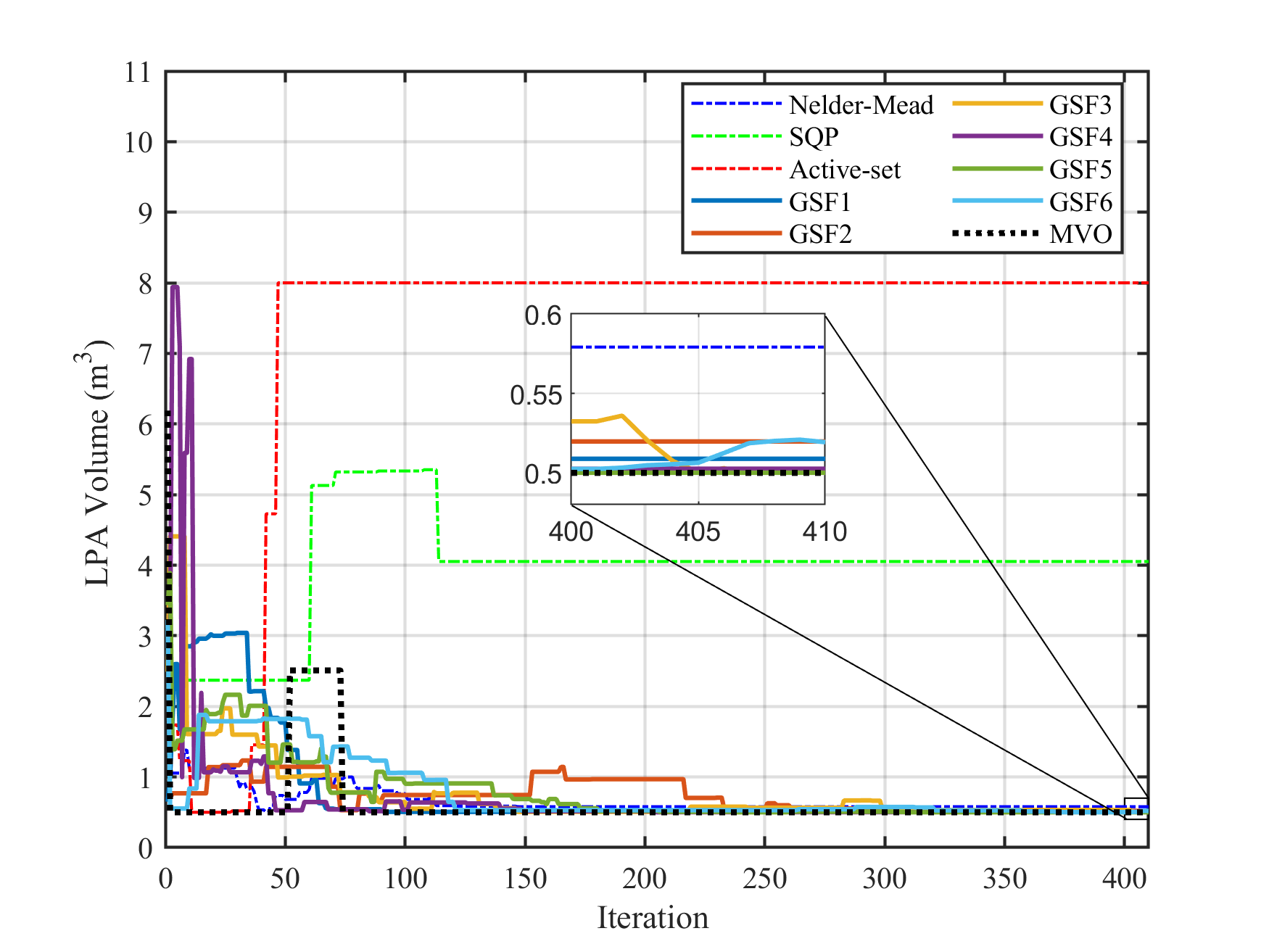}
    \caption{Convergence curve of the second decision variable, the LPA volume, for all the 10 optimization algorithms (6 GSF algorithms, 3 numerical algorithms, and MVO)}
       \label{CC_VL0}
\end{figure}

The next studied parameter is the low-pressure gas accumulator, which its chronological variation for the 10 optimization algorithms is illustrated in Fig. \ref{CC_VL0}.

This had the best converge out of all the 4 decision variables. As it is apparent in Fig. \ref{CC_VL0} eight out of 10 algorithms found the optimum value to be around 0.5 m$^3$ which is the minimum possible value for this parameter. The remaining algorithms are SQP and Active-set. This was the only decision variable that these 2 algorithms did not find a close optimum value for. SQP and Active-set converged on a value of 4 and 7 m$^3$ respectively, which cover over half of the feasible area.

\begin{figure}[H]
    \centering
    \captionsetup{justification=centering}
    \includegraphics[width=1\linewidth]{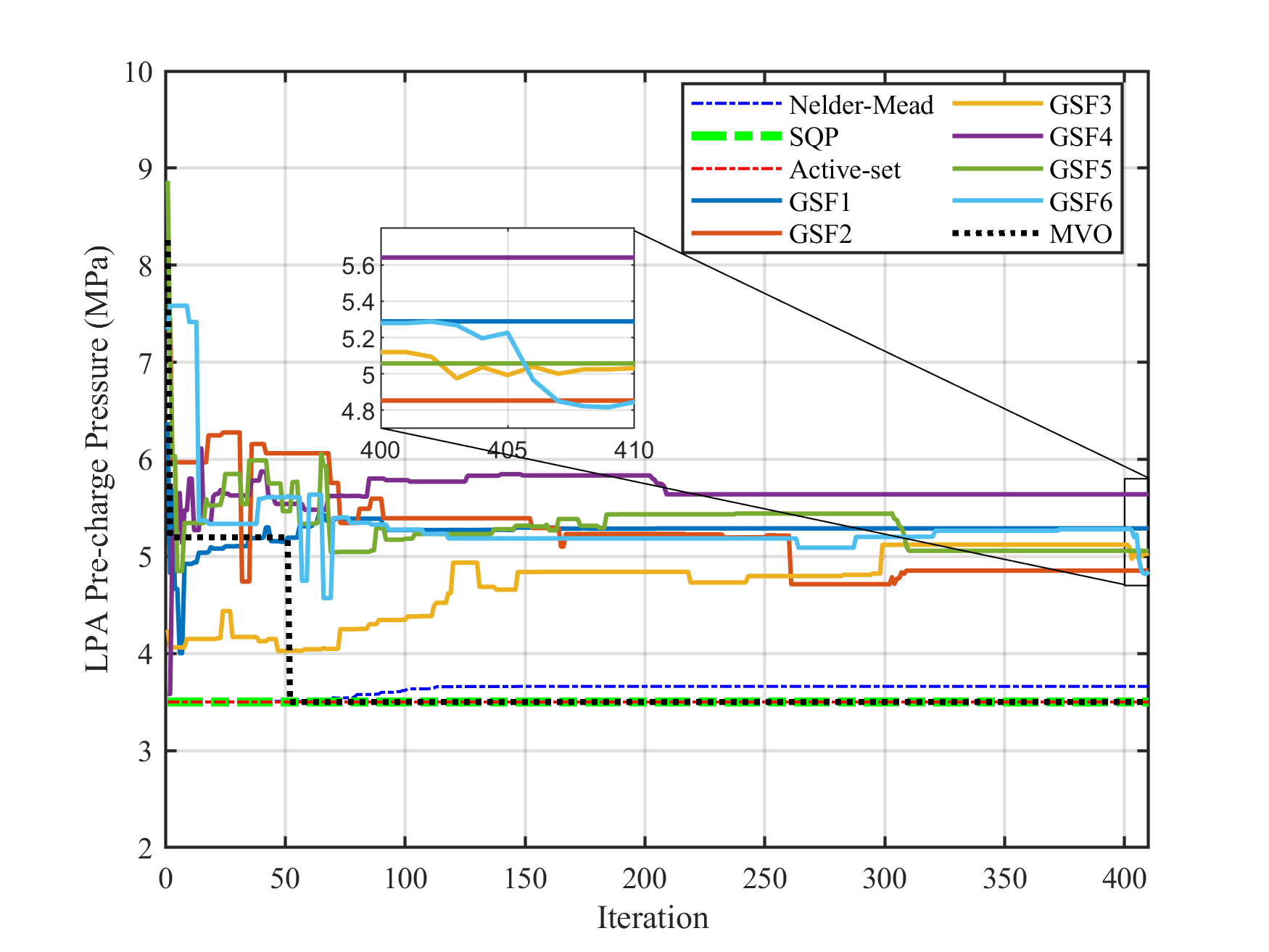}
    \caption{Convergence curve of the second decision variable, the LPA pre-charge pressure, for all the 10 optimization algorithms (6 GSF algorithms, 3 numerical algorithms, and MVO)}
       \label{CC_PL0}
\end{figure}

The last figure in this section is Fig. \ref{CC_PL0} which illustrates the convergence curve of all 10 algorithms for the low-pressure gas accumulator pre-charge pressure.

Based on Fig. \ref{CC_PL0}, the GSF algorithms were able to agree on a relatively close value for this parameter, which is around 4.8 to 5.6 MPa. The other 4 algorithms including the MVO and the numerical algorithms found the optimum value near the minimum range of the search space, almost a 3.5 MPa pre-charge pressure as the optimum value.

In this section, we aim to compare the performance of an RM3 heaving device with an HPTO unit with two sets of parameters. One uses the parameters of the Non-Optimized Case, which was extracted from WEC-Sim’s default case; the other uses the parameters of the best solution found by the 10 optimization algorithms, which belonged to the GSF2, with a 410 total number of iterations. The comparison is made by investigating the performance of different parts of the HPTO unit and plotting parameters like the pressures of the bottom and top chambers of the hydraulic cylinders, the PTO force applied to the WEC device, the hydraulic motor speed, the pressure differential between the HP- and LP-accumulators, hydraulic motor volume, the generator damping, and power throughout the simulation.

\begin{figure}[H]
    \centering
    \captionsetup{justification=centering}
    \includegraphics[width=1\linewidth]{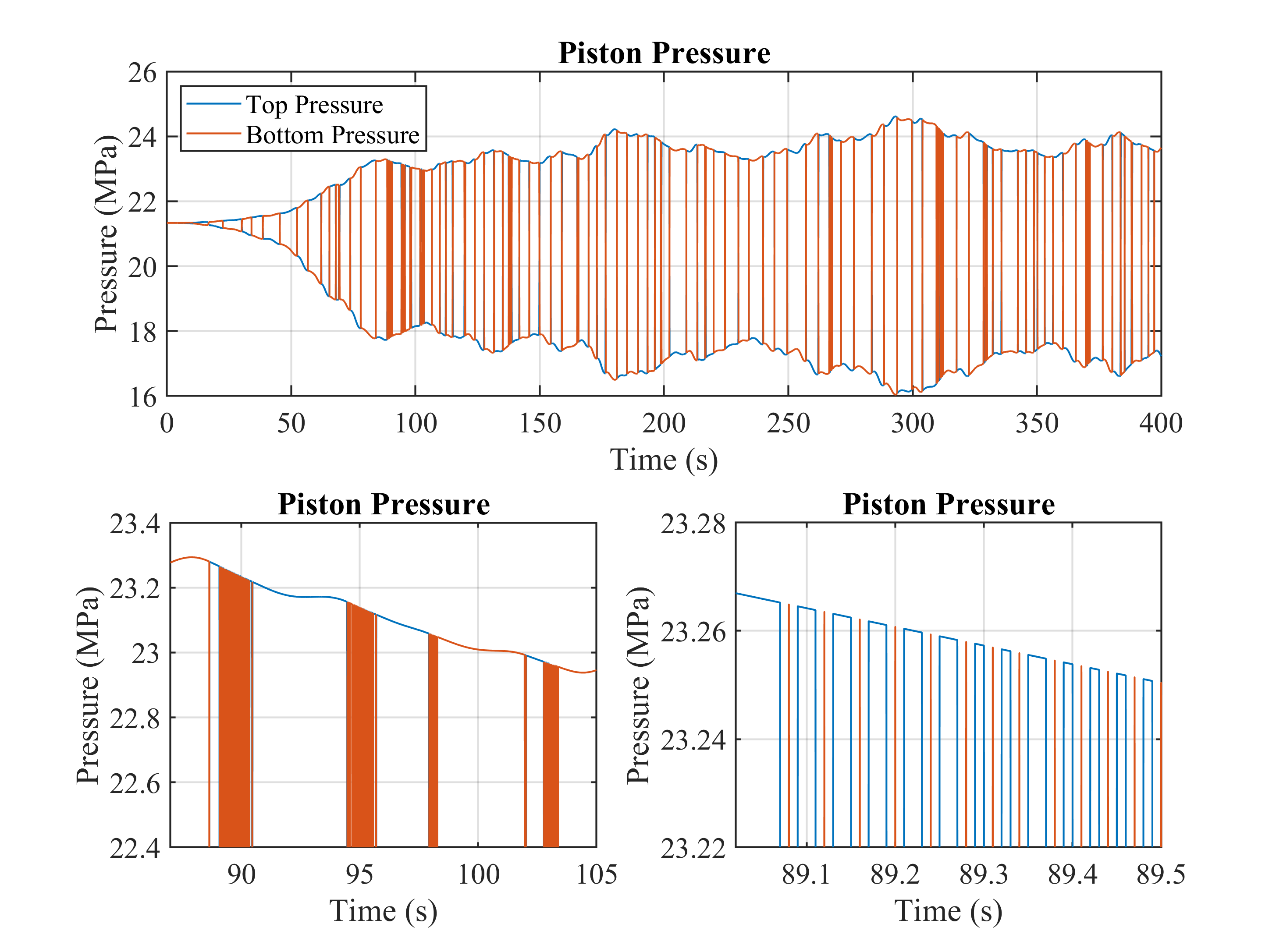}
    \caption{Hydraulic Cylinder Chambers Pressures during the simulation for the Non-Optimized Case}
       \label{Case0_Pic1}
\end{figure}

The first part of the hydraulic PTO is the hydraulic cylinder. We will look into the pressure of both pump chambers, which is illustrated in Fig. \ref{Case0_Pic1}. 

The value of top and bottom piston pressures vary from 16.1 to 24.6 MPa. It also has multiple fast fluctuation periods during the simulation time. Piston pressures reach their maximum and minimum value around the 293 s mark.

\begin{figure}[H]
    \centering
    \captionsetup{justification=centering}
    \includegraphics[width=1\linewidth]{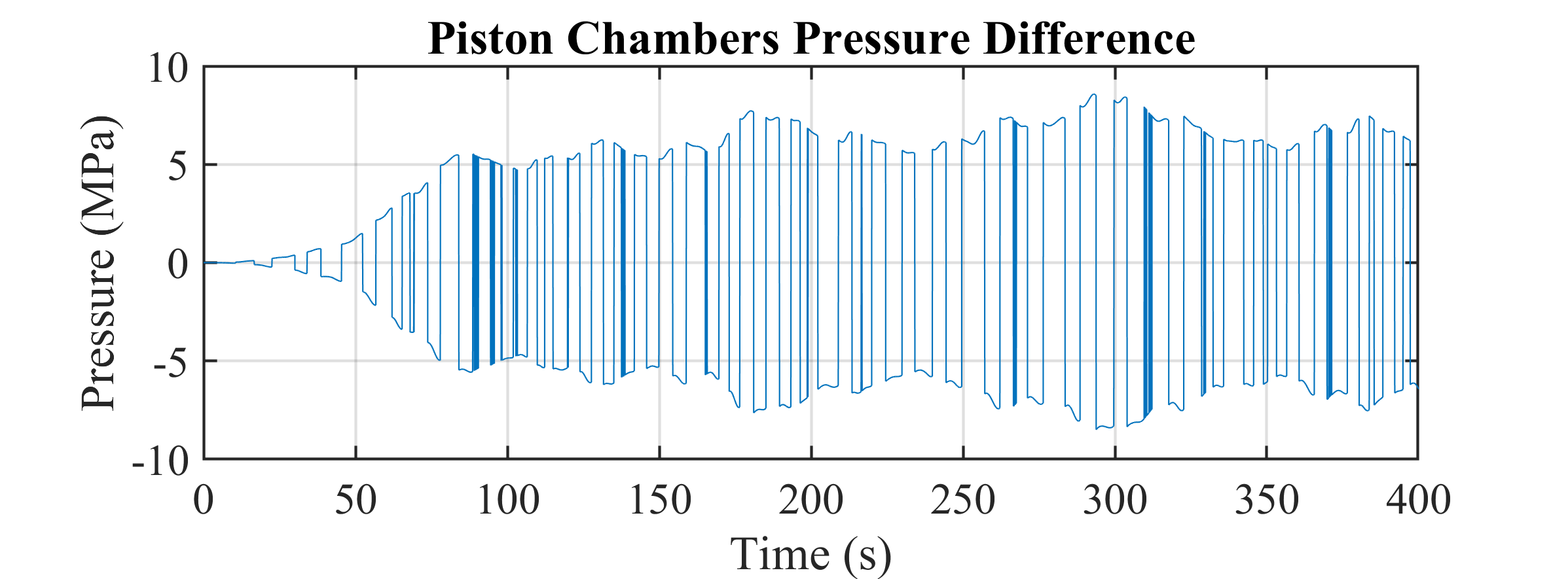}
    \caption{Pressure Difference Between the Hydraulic Cylinder Chambers during the simulation for the Non-Optimized Case}
       \label{Case0_Pic2}
\end{figure}

\begin{figure}[H]
    \centering
    \captionsetup{justification=centering}
    \includegraphics[width=1\linewidth]{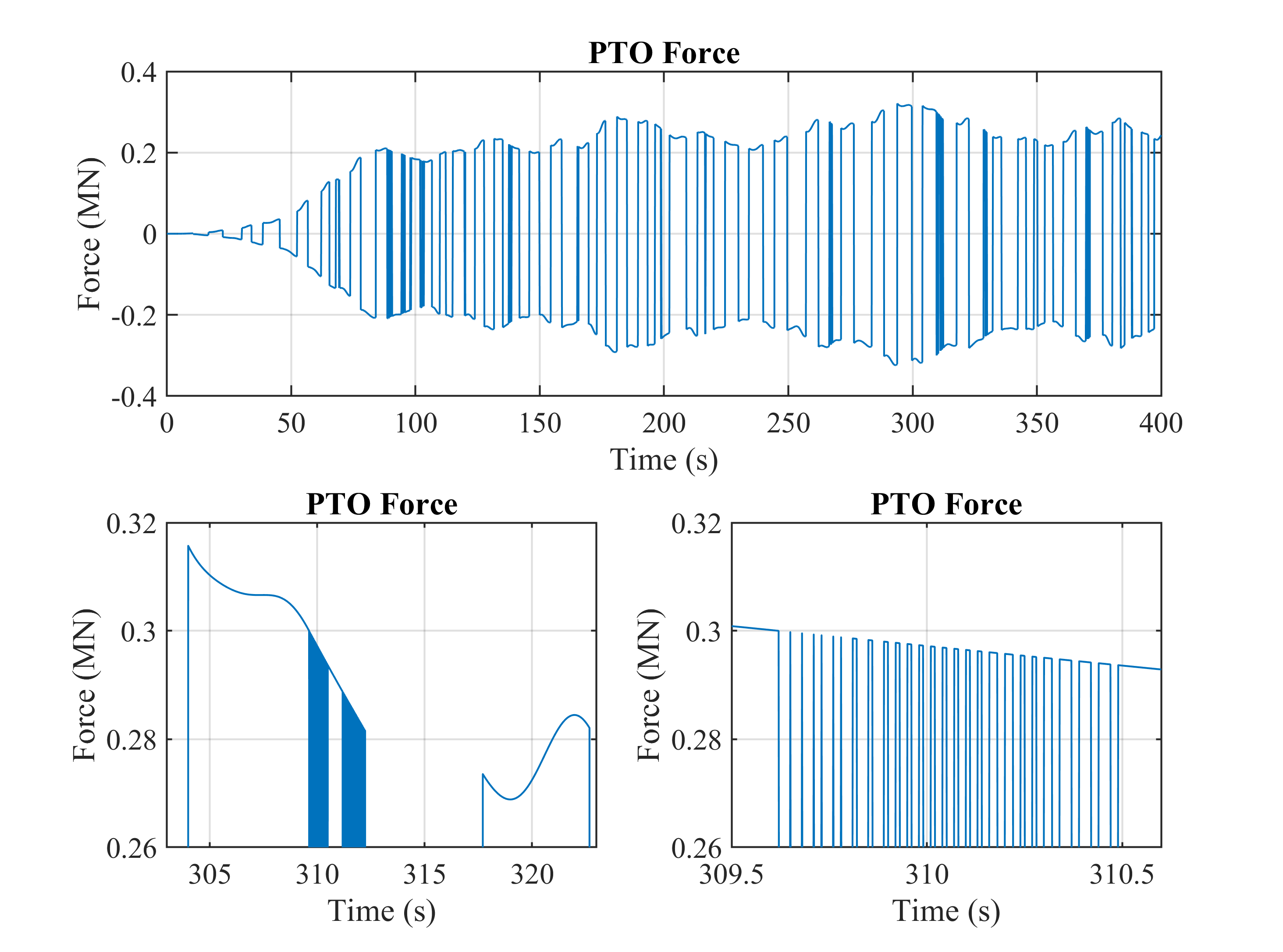}
    \caption{PTO Force during the simulation for the Non-Optimized Case}
       \label{Case0_Pic3}
\end{figure}

According to equation \ref{FPTO}, the PTO force value is dependant on the pressure difference between the upper and lower sections of the hydraulic piston; we are plotting the chronological change of these two parameters in Figs \ref{Case0_Pic2} and \ref{Case0_Pic3}.

Since the $A_p$ is constant through a simulation, the value of piston chambers pressure difference and the PTO force have a linear correlation. These two plots have similar overall shapes as expected. The maximum piston pressure difference value is 8.6 MPa which corresponds to the trough PTO force (-0.324 MN) at 293.1 s, and the minimum piston pressure difference value is -8.5 MPa which results in the peak PTO force (0.321), at 293.82 s.

\begin{figure}[H]
    \centering
    \captionsetup{justification=centering}
    \includegraphics[width=1\linewidth]{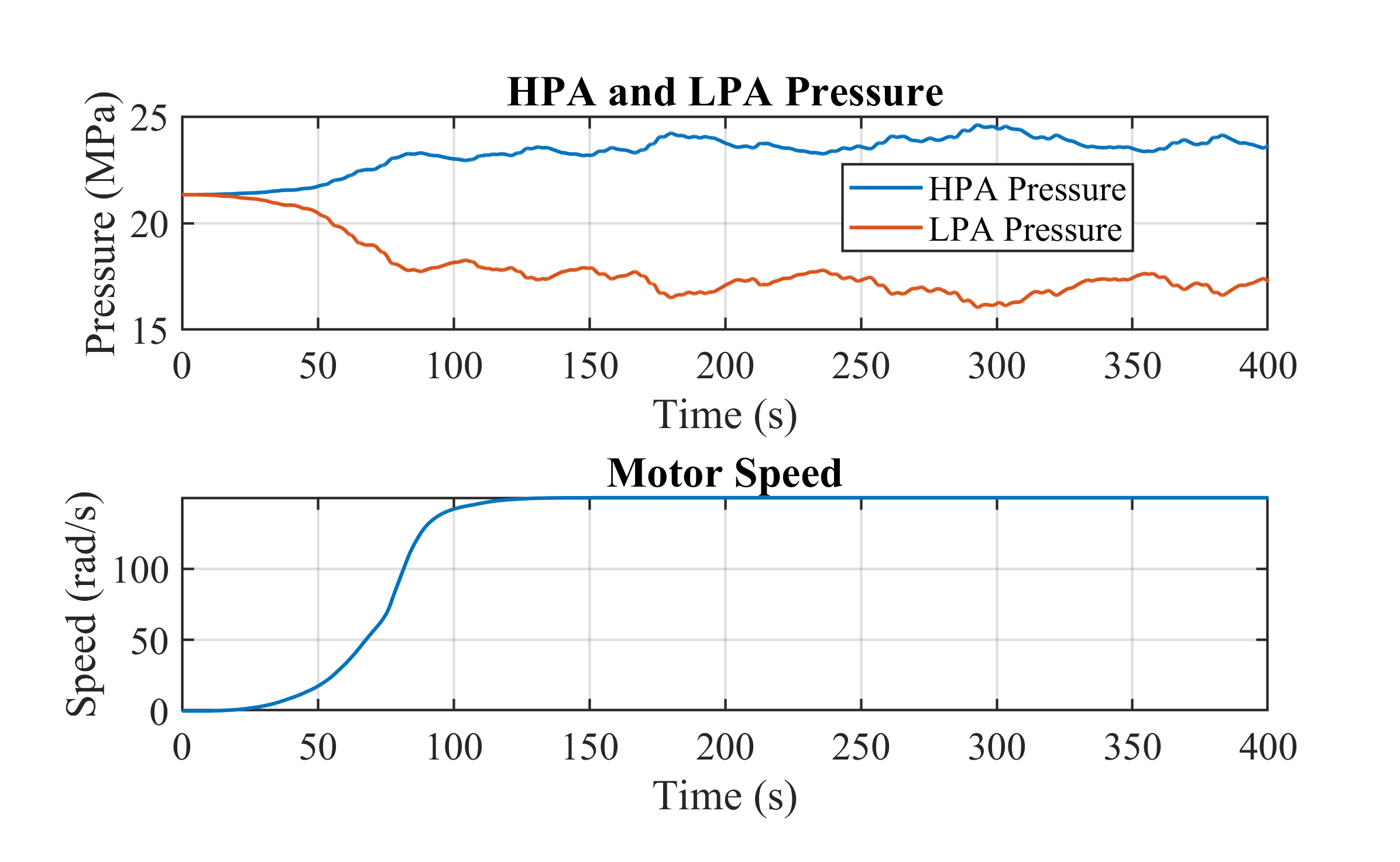}
    \caption{HP- and LP-accumulators Pressures and Hydraulic Motor Speed during the simulation for the Non-Optimized Case}
       \label{Case0_Pic4}
\end{figure}

Next, the HPA and LPA pressures and the hydraulic motor angular velocity are shown in Fig. \ref{Case0_Pic4}. The accumulators' job is to maintain an approximate constant $\Delta P$  across the hydraulic motor, resulting in constant motor angular speed and steady energy \cite{cargo2012determination}.

According to Fig. \ref{Case0_Pic4}, in the early stage of simulation, after around the 120 s mark, the HPA and LPA pressures become relatively stable. The maximum HPA pressure is 24.62 MPa, and the minimum LPA pressure is 16.04 MPa, which happens simultaneously at 293.1 s. Also, in the same figure, we can see that up until the 138 s, the motor speed is ramping up to reach the desired value, which is 150 rad/s. At this stage, the volumetric efficiency becomes 100 \%.

\begin{figure}[H]
    \centering
    \captionsetup{justification=centering}
    \includegraphics[width=1\linewidth]{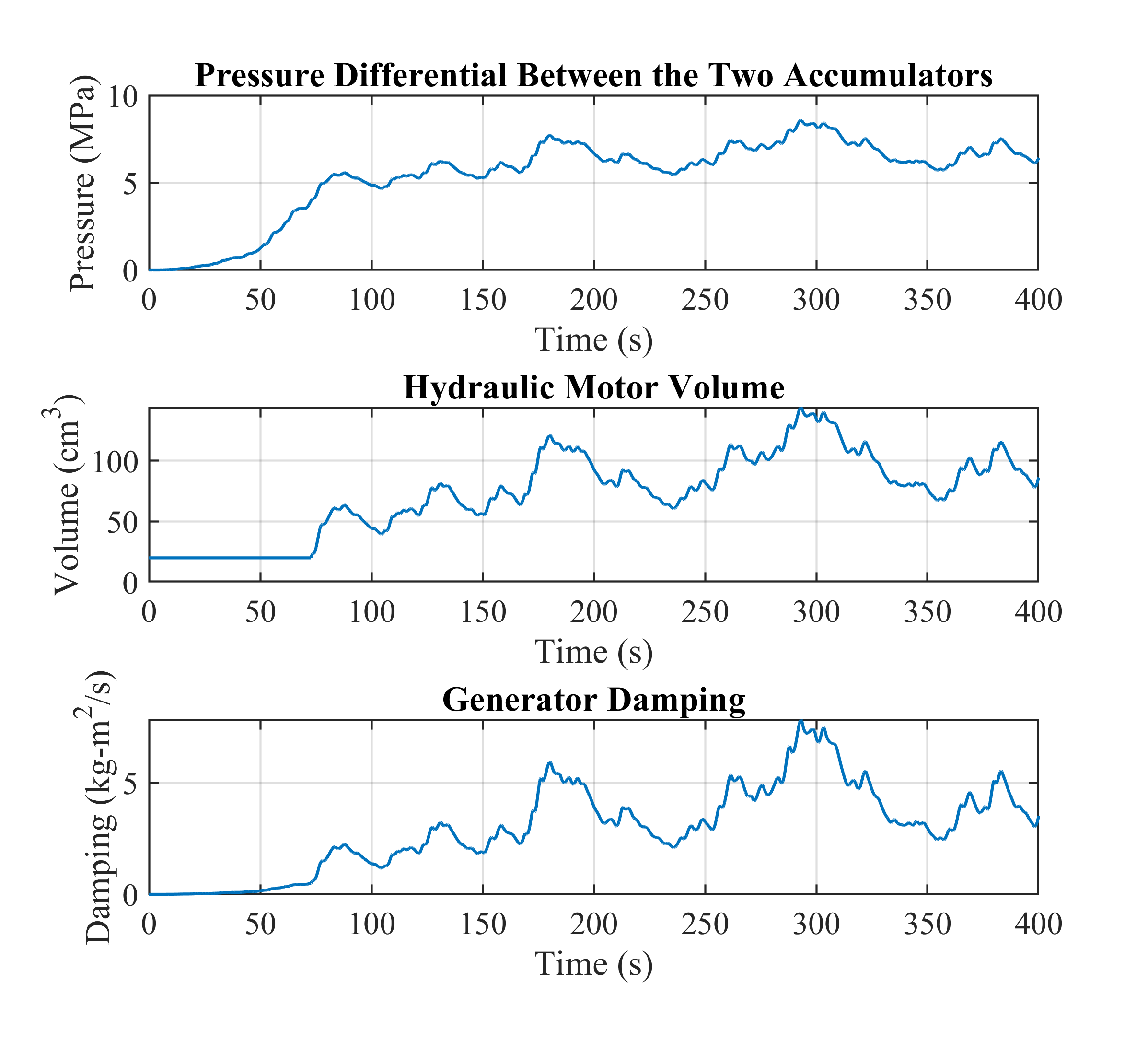}
    \caption{Pressure Differential Between the HP- and LP- Accumulators, the Hydraulic Motor Volume, and the Generator Damping during the simulation for the Non-Optimized Case}
       \label{Case0_Pic5}
\end{figure}

The following analyzed parameters are the accumulators’ pressure difference, the motor’s volume, and the generator damping, illustrated in Fig. \ref{Case0_Pic5}.

The relation between the parameters shown in Fig. \ref{Case0_Pic5}, plus the motor speed in Fig. \ref{Case0_Pic4}, can be determined from Equation \ref{Eq:Cgen}. All three plots in Fig. \ref{Case0_Pic5} have similar trends, but it can be seen that the standard deviation for the first parameter is lower than the other two. For example, although there is a mild descending trend in accumulators’ pressure difference between 215 s and 236 s, the more steep trend in this interval for the hydraulic motor volume resulted in a larger generator damping drop during these 21 seconds. Notably, the maximum value for all these three parameters occurs at 293.1 s, and they are equal to 8.6 MPa, 143.8 cm$^3$, and 7.8 kg.m$^2$/s. 

Although $\Delta$P reaches its average value after 75 s and oscillate around that, because motor displacements ascending trend did not start until around the 100 s mark, the 100 s value is considered as the model’s ramp time to calculate the parameters, including the generator’s damping, after this point in time.

\begin{figure}[H]
    \centering
    \captionsetup{justification=centering}
    \includegraphics[width=1\linewidth]{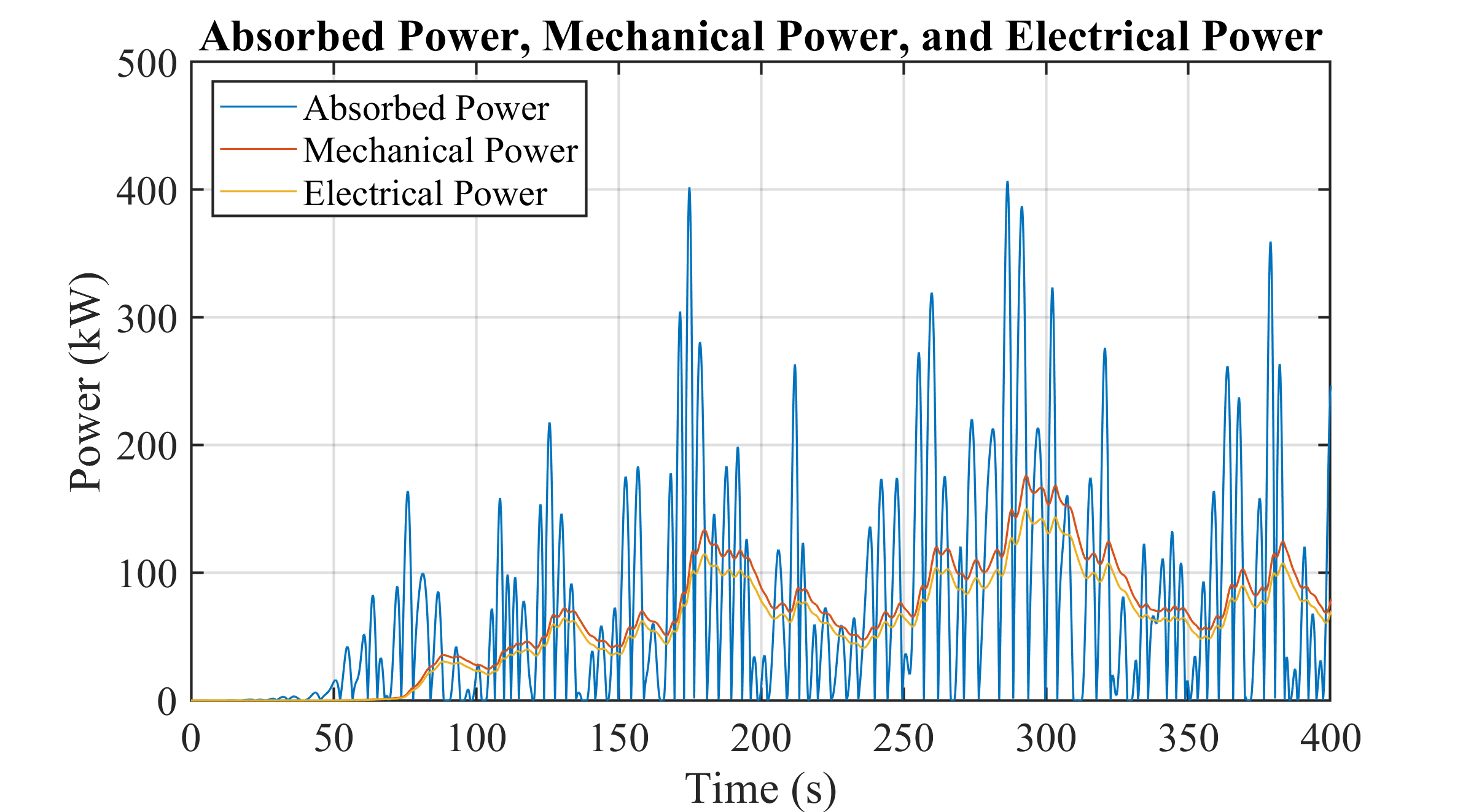}
    \caption{Absorbed Power, Mechanical Power, and Electrical Power during the simulation for the Non-Optimized Case}
       \label{Case0_Pic6}
\end{figure}

Now, we will take a closer look at the power generation parameters. The absorbed, mechanical and electrical powers are depicted in Fig. \ref{Case0_Pic6}.

As you can see, the absorbed power has multiple fluctuations at small time intervals, but the other two powers have a more smooth profile, which is the result of using accumulators. The maximum absorbed power is 407 kW which happens at 286.5 s. Also, the electrical power is lower than the mechanical power throughout the simulation, which is the result of losses in the generator.

Now we will study the second case, i.e., the best solution found by the GSF2 algorithm. This will be done by plotting the same plots and comparing them with their correspondents from the non-optimized case. 

\begin{figure}[H]
    \centering
    \captionsetup{justification=centering}
    \includegraphics[width=1\linewidth]{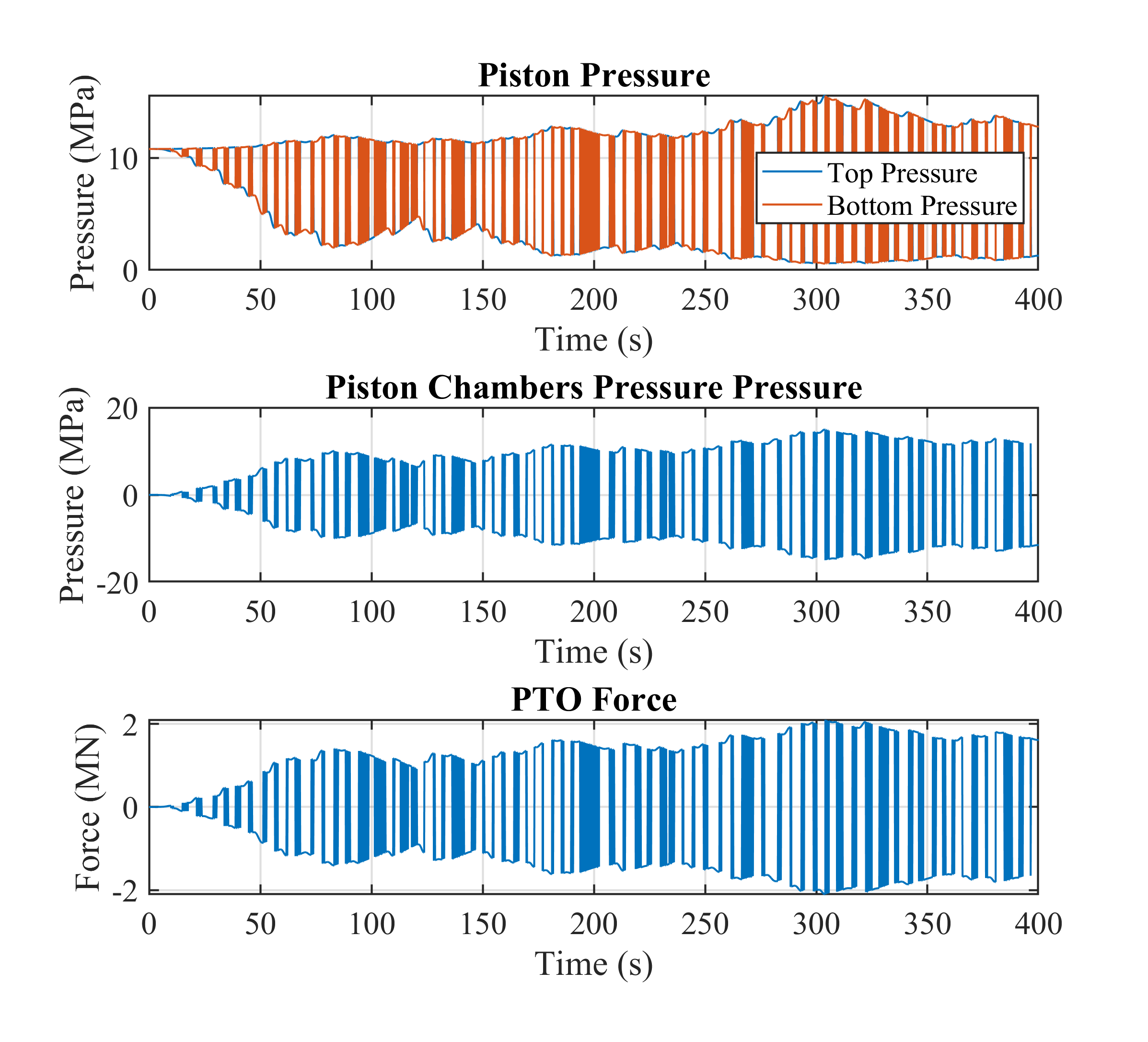}
    \caption{Hydraulic Cylinder Top and Bottom Chambers Pressures, and Chambers Pressure Difference, and PTO Force during the simulation for the Best Case Found by the Optimization Algorithms (GSF2)}
       \label{Case5_Pic123}
\end{figure}

Fig. \ref{Case5_Pic123} shows the chronological alteration of piston chambers pressure, the pressure difference between the two piston chambers, and the PTO force; these values are connected by Equation \ref{FPTO}.

According to Fig. \ref{Case5_Pic123}, the piston upper and lower section pressures have multiple intervals with high-frequency fluctuations. The maximum and minimum values for these parameters are 15.5 MPa and 0.5 MPa, which happen at the 304 s mark. Compared to the non-optimized case, the piston pressures have the same range but are significantly lower and have no shared values.

Also, the piston pressure difference values have the same high-frequency fluctuations intervals between -14.975 MPa and 14.995 MPa; these extreme values are almost the ones for the non-optimized case, and they happen 304 seconds after the simulation started.

In the last plot of Fig. \ref{Case5_Pic123}, one can see the PTO force throughout the simulation, which according to Equation \ref{FPTO}, is directly related to the piston pressure difference. At 304 s, we can see the ultimate values -2.103 and 2.100 MN, almost 7 times the minimum and maximum PTO force values for the non-optimized case. The increase of PTO force is by 556 \% prevents the converter from operating during the smaller wave condition \cite{jusoh2020parameters}.

\begin{figure}[H]
    \centering
    \captionsetup{justification=centering}
    \includegraphics[width=1\linewidth]{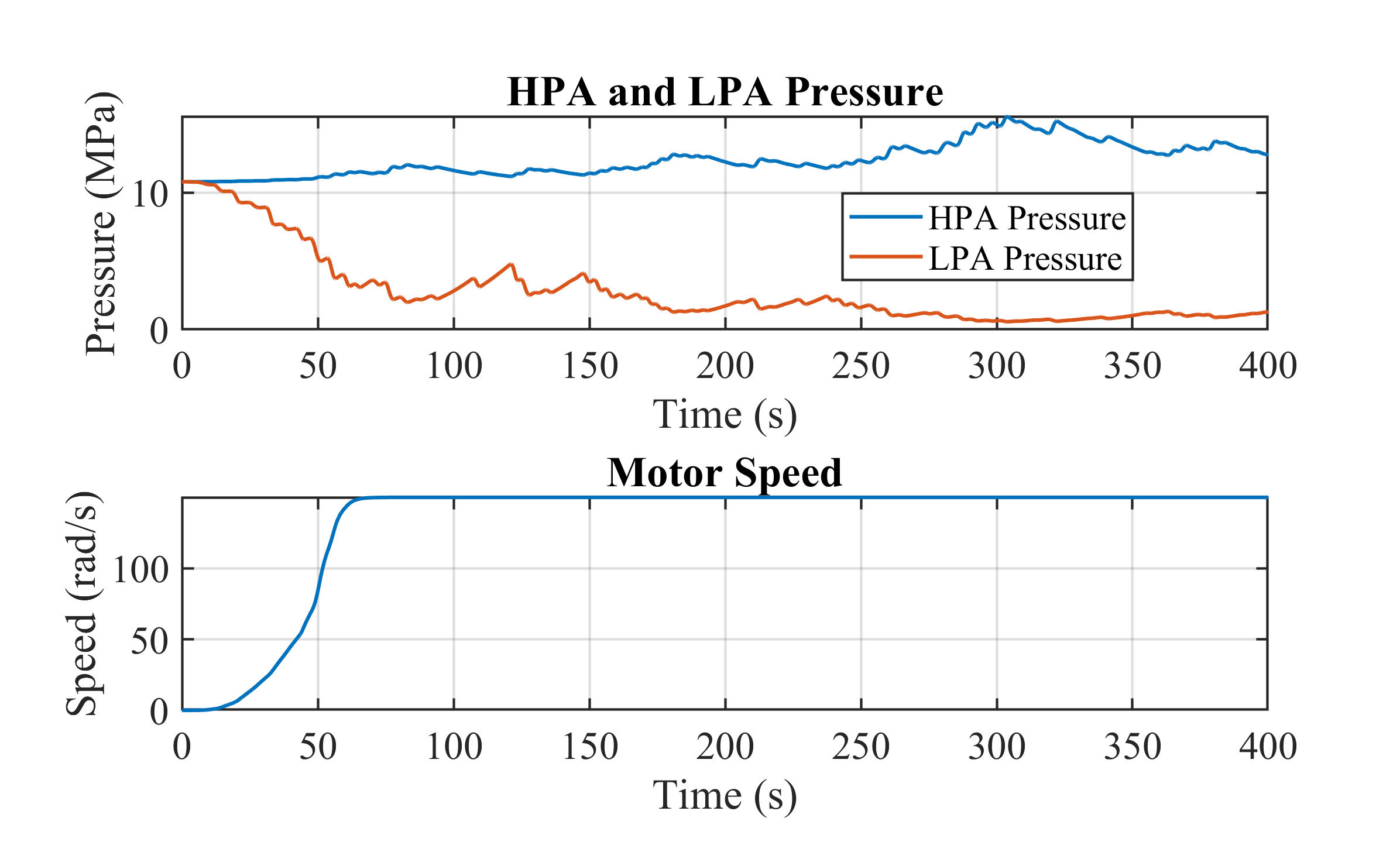}
    \caption{HP- and LP-accumulators Pressures and Hydraulic Motor Speed during the simulation for the Best Case Found by the Optimization Algorithms (GSF2)}
       \label{Case5_Pic4}
\end{figure}

According to Fig. \ref{Case5_Pic4}, which presents the HPA and LPA pressure and the hydraulic motor angular velocity for the GSF2 case, we can see that the HPA pressure is steady at the first half of the simulation and the LPA pressure is stable in the other half. Also, the range of HPA pressure variation is not the same as the LPA pressure’s one, but for the non-optimized case, they were almost the same, and also, their profile was virtually symmetrical. Also, the motor speed reaches its desired value at 77.8 s, which is nearly half the required time for the non-optimized case.

\begin{figure}[H]
    \centering
    \captionsetup{justification=centering}
    \includegraphics[width=1\linewidth]{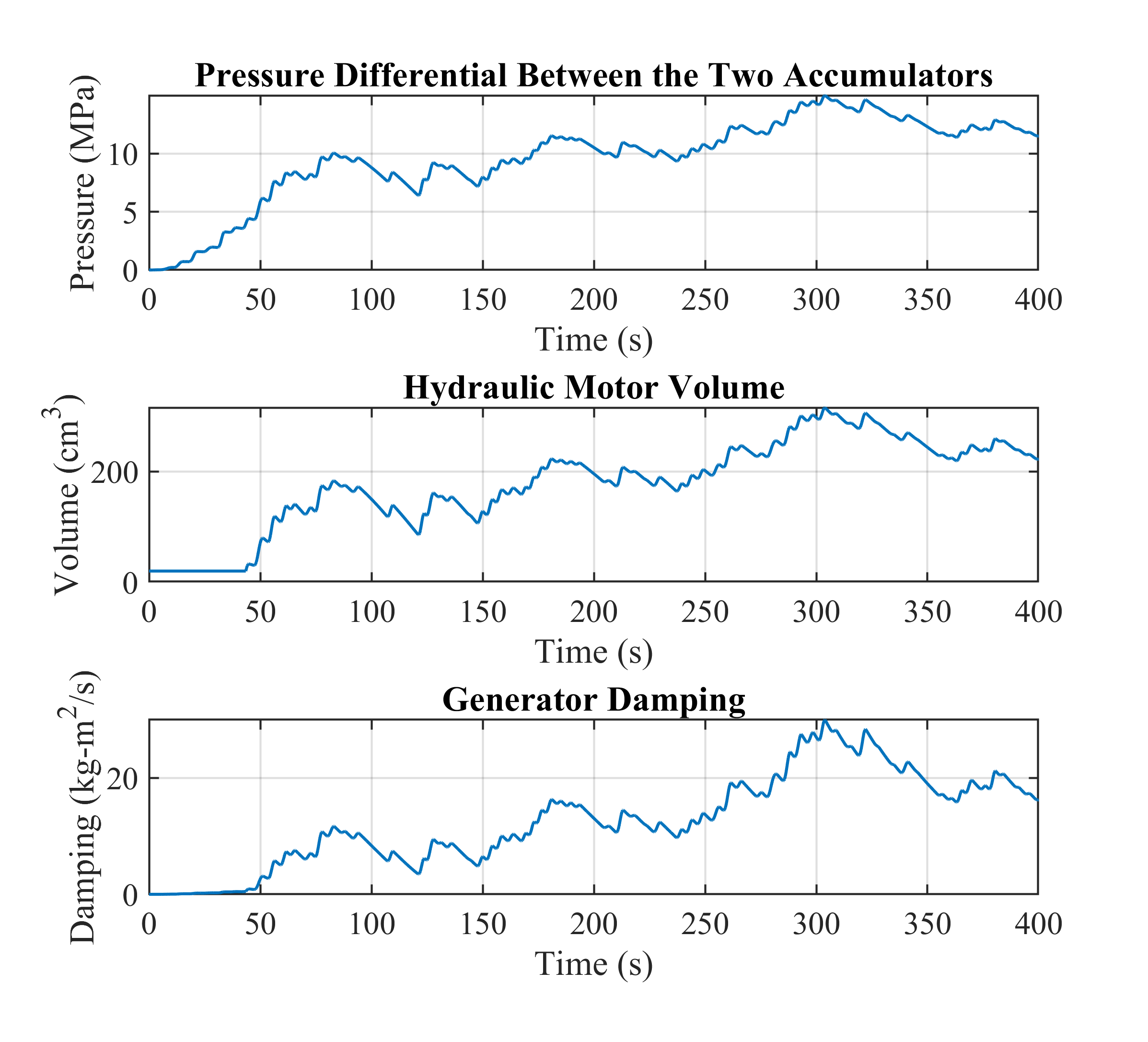}
    \caption{Pressure Differential Between the HP- and LP- Accumulators, the Hydraulic Motor Volume, and the Generator Damping during the simulation for the Best Case Found by the Optimization Algorithms (GSF2)}
       \label{Case5_Pic5}
\end{figure}

Fig. \ref{Case5_Pic5} shows the accumulators' pressure difference, the motor volume, and the generator damping for the best solution found by the optimization algorithms. Again, these parameters are related to each other based on Equation \ref{Eq:Cgen}.

The maximum values for these three parameters occur at 303.84 s when the accumulator’s pressure difference is 14.995 MPa, and the hydraulic motor volume is 315.17 cm$^3$; finally, the generator damping is 30 km.m$^2$/s. These values are almost 2, 2, and 4 times the non-optimized case values.

\begin{figure}[H]
    \centering
    \captionsetup{justification=centering}
    \includegraphics[width=1\linewidth]{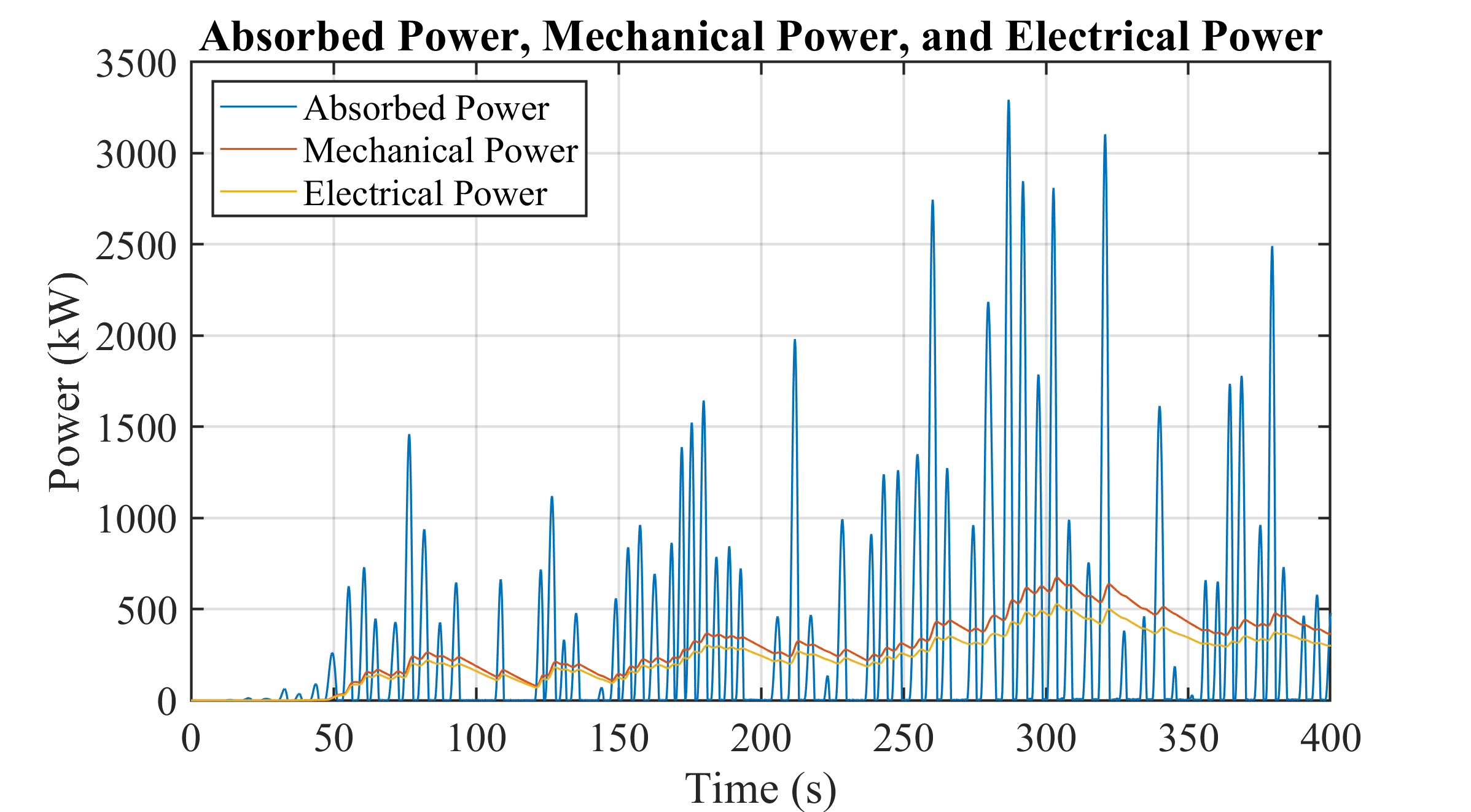}
    \caption{Absorbed Power, Mechanical Power, and Electrical Power during the simulation for the Best Case Found by the Optimization Algorithms (GSF2)}
       \label{Case5_Pic6}
\end{figure}

And finally, for the power generation of the HPTO, we plotted the absorbed, generated, and electrical powers for the best-optimized case in Fig. \ref{Case5_Pic6}. Similar to Fig. \ref{Case0_Pic6}, we can see that the absorbed power has high-frequency fluctuations and generated and electrical power has more smooth profiles, which is more smooth than the Fig. \ref{Case0_Pic6} plots. The reason can be using an HPA with a higher volume in this case, despite the increase in the LPA volume \cite{jusoh2021estimation}. Decreasing the HPA and LPA volumes nullifies the smoothing effect and makes the electrical and mechanical powers behave more like the absorbed power \cite{quartier2018numerical}. The maximum absorbed power is 3293 kW which is 7 times more than the non-optimized case value, and it happens at 286.92 s.

It is notable that since the capacity of the HPA in our optimized case has increased, it results in an increase in size, weight and cost of the HPTO unit, since the hydraulic accumulator and the hydraulic motor are the most expensive parts of the HPTO system \cite{jusoh2021estimation}.

\section{Conclusions} 
\label{Conclusion}
In the study, the configuration parameters of a hydraulic PTO connected to an RM3 point absorber have been optimized to reach the maximum power output. Ten optimization has been used for this purpose, including three numerical algorithms, MVO, and six Modified Combined Methods. The simulation–optimization of HPTO unit parameters has been carried out using WEC-Sim in MATLAB software. 
We did a sensitivity analysis to find out the effect configuration parameters on the power output and the power fluctuation ratio ($R_{PF}$), and it showed that:
\begin{itemize}
  \item Low values of $A_p$ result in low power output and low $R_{PF}$.
  \item Low values of $LPA$ pre-charge pressure result in good average power output and low $R_{PF}$. 
  \item Small and moderate initial volumes of the HPA can increase the 
  \item No distinct pattern was found for the influence of LPA volume change on the power output and the $R_{PF}$.
\end{itemize}
The modified Genetic, Surrogate, and fminsearch algorithm combinations had the most promising results and were able to increase the mean power output by 304 \% and decrease the $R_{PF}$ by 13 \% by adjusting the HPTO parameters to their optimal values. Good convergence of the results of this algorithm is obtained by comparing six different optimisations, which generally achieve a parameter convergence range of less than 3\% of the initial range. The most convergent parameters are the high and low pressure gas accumulator volumes, which converge to the maximum and minimum values of the considered range respectively. GSF2 and GSF6 produced the best results and an optimal power around 230kW. \\
Future work could also consider the effect of the cost of the HPTO on the optimal design of the system. For example, a larger HPA volume means an increase in the size, weight, and cost of the HPTO system.
Furthermore, the simultaneous interaction of hydraulic transmission system specifications over the change of incident waves' frequency would be promising.


\vspace{6pt} 


\textbf{CRediT authorship contribution statement:} \\
\textbf{Erfan Amini}: Conceptualization, Simulation, Methodology, Original draft preparation, writing. \textbf{Hossein Mehdiour}: Simulation, Writing Original draft, Methodology, Visualisation. \textbf{Emilio Faraggiana}: Simulation, Visualisation, Writing original draft. \textbf{Danial Golbaz}: writing original draft. \textbf{Sevda Mozaffari}: Edit-ing original draft. \textbf{Giovanni Bracco}: Supervision, Editing original draft. \textbf{Mehdi Neshat}: Conceptualization, Supervision, Editing original draft.



\section*{Acknowledgement}
The project's results have carried out using High Performance Computing Laboratory (HPCL) clusters of School of Civil Engineering at University of Tehran , and the HACTAR High Performance Computing (HPC) of Politecnico of Turin.\\
\\
\textbf{Funding:} This research received no external funding.\\
\\
\textbf{Conflicts of interest:} The authors declare no conflict of~interest.\\
\\
\textbf{abbreviations:} The following abbreviations are used in this manuscript:\\

\noindent 
\begin{tabular}{@{}ll}
WEC & Wave Energy Converter\\
WEC-Sim & Wave Energy Converter Simulator\\
PTO & Power Take-off \\
RM3 & Reference Model 3 \\
OpenFOAM & Open Source Field Operation and Manipulation\\
OWC & Oscillating Water Column\\
OSWEC & Oscillating Surge Wave Energy Converter\\

\end{tabular}




\bibliographystyle{unsrt} 
\bibliography{bibliography}

\end{document}